\documentclass{article}

    \usepackage[preprint]{neurips_2023}

\usepackage[utf8]{inputenc} 
\usepackage[T1]{fontenc}    
\usepackage{hyperref}       
\usepackage{url}            
\usepackage{booktabs}       
\usepackage{amsfonts}       
\usepackage{nicefrac}       
\usepackage{microtype}      
\usepackage{xcolor}         
\usepackage{adjustbox}
\usepackage{multirow}
\usepackage{graphicx}
\usepackage{pgfplots} 
\usepackage{wrapfig}
\usepackage{booktabs}
\usepackage{pifont}
\usepackage{algorithm}
\usepackage{algpseudocode}
\usepackage{enumitem}
\usetikzlibrary{patterns}
\usepackage{amsmath, amssymb}
\usepackage{cleveref} 
\usepackage{subcaption}
\usepackage{caption}
\usepackage{setspace}

\newcommand{\cmark}{\ding{51}}%
\newcommand{\xmark}{\ding{55}}%

\newcommand{\snact}{ToolBench}

\title{On the Tool Manipulation Capability of \\Open-source Large Language Models}

\author{%
  Qiantong Xu, Fenglu Hong, Bo Li, Changran Hu, Zhengyu Chen, Jian Zhang \\
  SambaNova Systems, Inc. \\
  Palo Alto, CA, USA \\
  \texttt{\{qiantong.xu,jian.zhang\}@sambanovasystems.com} \\
}

\begin{document}

\maketitle

\vspace{-1.5em}
\begin{abstract}
Recent studies on software tool manipulation with large language models (LLMs) mostly rely on closed model APIs. The industrial adoption of these models is substantially constrained due to the security and robustness risks in exposing information to closed LLM API services.
In this paper, we ask \emph{can we enhance open-source LLMs to be competitive to leading closed LLM APIs in tool manipulation, with practical amount of human supervision}. 
By analyzing common tool manipulation failures, we first demonstrate that open-source LLMs may require training with usage examples, in-context demonstration and generation style regulation to resolve failures. 
These insights motivate us to revisit classical methods in LLM literature, and demonstrate that we can adapt them as model alignment with programmatic data generation, system prompts and in-context demonstration retrievers to enhance open-source LLMs for tool manipulation.
To evaluate these techniques, we create \textit{\snact}\footnote{Available at \url{https://github.com/sambanova/toolbench}}, a tool manipulation benchmark consisting of diverse software tools for real-world tasks.
We demonstrate that our techniques can boost leading open-source LLMs by up to $90\%$ success rate, showing capabilities competitive to OpenAI GPT-4 in $4$ out of $8$ \snact\  tasks.
We show that such enhancement typically requires about one developer day to curate data for each tool, rendering a recipe with practical amount of human supervision.
\end{abstract}

\section{Introduction}
\label{sec:intro}

Tool-augmented large language models (LLMs) recently emerge as a research frontier. Such augmented LLMs demonstrate tool manipulation capabilities which automate software operations through natural language instructions~\cite{schick2023toolformer,li2023api,qin2023tool,autogpt,shen2023hugginggpt}. 
Despite the fact that open-source LLMs substantially shrink the quality gap towards proprietary closed LLMs in tasks such as chatbot~\cite{vicuna2023,openchatkit,laionOA,dolly2dot0}, recent tool-augmented LLMs still mostly rely on closed LLM APIs~\cite{schick2023toolformer,li2023api,qin2023tool,autogpt}. 
This leads to a fundamental barrier for the industrial adoption of these augmented LLMs due to security and robustness risks associated with exposing enterprise-internal workflows and information to closed LLM APIs~\cite{samsung,jpmorgan}.
To maximize the industrial impact, there is a substantial need for tool manipulation capabilities founded on open-source LLMs. 
To this end, we ask \emph{can we build on open-source LLMs with practical amount of human supervision and achieve tool manipulation capabilities competitive to closed LLMs.}

\begin{figure}[t!]
\caption{Tool manipulation setup. 
We augment LLMs as action generators with access to API documentations. 
In a single-step scenario, an action generator directly generates API calls to accomplish the goal. A multi-step action generator iterates with an environment using API calls and generates the next-step calls based on the information from the environment until an exit state.
}
\centering
\includegraphics[width=\textwidth]{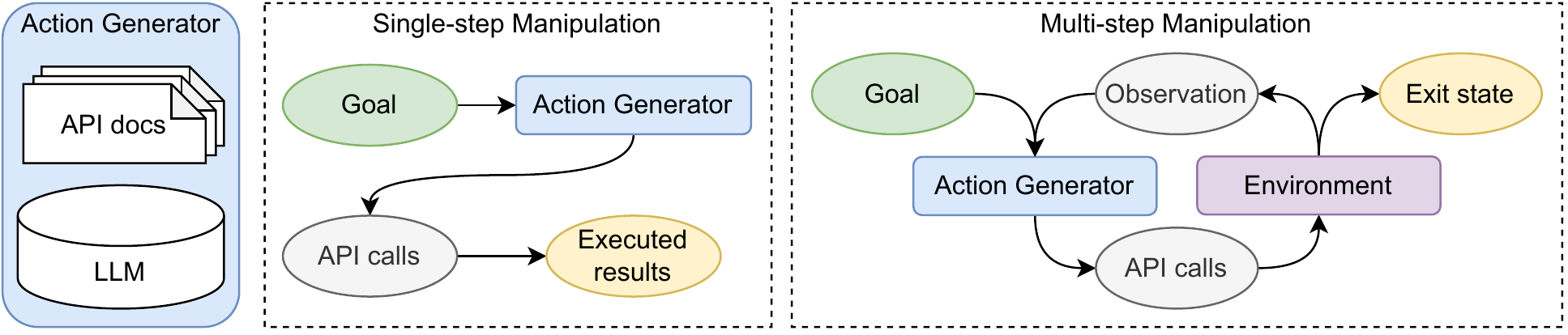}
\label{fig:task_setup}
\vspace{-1em}
\end{figure}

In this paper, we first demystify key challenges for tool manipulation using open-source LLMs; we then leverage the insights to suggest practical recipes for enhancement. 
Concretely, we study the setting shown in~\Cref{fig:task_setup} where LLMs take in a natural language instruction as the goal and generate API calls to accomplish the goal. 
Although we expect a quality gap between the open-source and closed LLMs~\cite{liang2022holistic}, what we observe is a far more severe disparity. Specifically, for an on-sale house searching tool, a leading open LLM for code generation fails every test case while the OpenAI GPT-4~\cite{OpenAI2023-ov} attains $77\%$ success rate across the same one hundred examples. 
This observation motivates us to study the challenges for open-source LLMs to attain strong tool manipulation capability.

During our investigation, we identify three key challenges that impede the performance of open-source LLMs in tool manipulation. 
Firstly, open-source models often struggle to accurately identify API names, whereas closed LLMs demonstrate the capability to invoke the correct APIs without explicit usage examples or documentation during inference. 
This suggests that closed LLMs hypothetically internalize knowledge of API usage during training. 
Secondly, we show that without demonstration examples, open-source LLMs often fail to populate the appropriate values for API arguments. 
Thirdly, we demonstrate that open-source LLMs tend to produce non-executable generation, such as natural language beyond the desired code. 

Our insights suggest us to revisit three \emph{simple} techniques from LLMs for conventional NLP tasks.
In the context of tool manipulation, 
we adapt them with practical amount of supervision and use them to enhance open-source LLMs.
\emph{Model alignment:} To first internalize API usage knowledge, we perform instruction tuning~\cite{ouyang2022training,bai2022constitutional} with programatically generated data. Specifically, we first write a few dozens of templates on goals and corresponding API calls. We then pragmatically bootstrap the data volume by instantiating templates with concrete key word values.
\emph{In-context demonstration retriever:} Inspired by retrieval-augmented generation~\cite{borgeaud2022improving,li2022survey,ram2023context}, we additionally enhance the LLMs with a \emph{retriever} to leverage in-context demonstrations during inference. This module selects demonstration examples with the most semantically similar goals from a human-curated pool of examples. Given $n$ API functions, the retriever only requires $\mathcal{O}\left(n\right)$ examples where every API function appears in at least one example. We then leverage LLMs to generalize to goals achieved by unseen API combinations. 
\emph{System prompt:} Finally we embed goal descriptions into a pre-defined system prompt which provides inference-time guidelines to  generate executable API calls; such system prompts were shown to regulate language style in chatbots~\cite{glaese2022improving}. 
These techniques only require a small amount of human supervision. Thus they render a potentially practical recipe for building on top of open-source LLMs.

To extensively evaluate the inspired techniques, we present \emph{\snact}, a benchmark suite on eight diverse tools ranging from Google Sheets manipulation to controlling robots~\cite{liang2022code}. It enables the first publicly-available quantitative evaluation test bench among the ones brought up in the tool-augmented LLM literature \cite{li2023api, qin2023tool}. For the software tools in our benchmark, LLMs need to accomplish a variety of goals by selecting and combining API functions from up to a hundred candidates. 

Using the tools in the \snact\  suite, we first empirically show that leading open-source LLMs can demonstrate up to $78\%$ lower success rate when compared to the OpenAI GPT-4 APIs.
We then demonstrate that these simple techniques can substantially improve the success rate of open-source LLMs by up to $90\%$, attaining results competitive or better than OpenAI GPT-4 models in $4$ out of the $8$ tools in our benchmark\footnote{We apply the same system prompt and in-context example retriever for GPT-4. Model alignment is not applicable to GPT-4 as there is no publicly available tuning APIs for it during our experiments.}.
To reveal the impact of different techniques, we provide evidence that aligning model with synthetic data primarily contributes to the significant improvement of open-source LLMs. The system prompt and the in-context demonstration retriever further enhance the performance.
During the enhancement process, we observe that, on average, it takes just one day for a developer to craft the in-context demonstrations and curate the templates for generating model alignment data. This implies that the recipe requires a practical level of human supervision. 


Our contributions and the structure of this paper are as follows.
\begin{itemize}[leftmargin=*]
\item In~\Cref{sec:insights}, we reveal challenges in API selection, argument populating and non-executable generation which hinder open-source LLMs on tool manipulation.
\item To alleviate the challenges, we revisit simple techniques for conventional NLP tasks. We adapt them for tool manipulation to boost open-source LLMs with minimal human supervision in~\Cref{sec:techniques}.
\item In~\Cref{sec:benchmarks}, we introduce the \snact, the first open-sourced benchmark with pre-defined test cases for quantitative evaluation compared to the ones in the recent tool-augmented LLM literature. 
\item We demonstrate in~\Cref{sec:experiments} that our adapted techniques boost open-source LLMs by up to $90\%$ success rate, showing competitiveness with GPT-4 APIs in $4$ out of $8$ \snact\  tasks.
\end{itemize}
\section{Background}
\label{sec:background}
To establish backgrounds, we first concretize the software tool manipulation setup. We then present a preliminary observation on the capability of open-source LLMs. This observation motivates our study on the challenges in~\Cref{sec:insights} which inspire simple techniques for enhancements in~\Cref{sec:techniques}.  

\subsection{Tool manipulation setup}
\label{subsec:tool_setup}
In this paper, we study the scenario where software users intend to translate a natural language goal description $\emph{g}$ into a sequence of application programming interface (API) calls $C_g = \left\{c_0, c_1, \cdots, c_{n_g} \right\}$ to accomplish the goal. We study tool manipulation with open-source LLMs in this specific setting, because APIs serve as the prevalent abstraction for developers and users in modern software systems. 

\paragraph{Large language model} Autoregressive language models encode probabilities of the next word $x_{N + 1}$ given $x_0, x_1, \cdots, x_N$ as the context sequence~\cite{dan_ngram}. By sampling from this conditional probability $p\left(x_{N + 1} | x_0, x_1, \cdots, x_N\right)$ iteratively, it generates language continuations from given contexts. In the recent wave of scaling up model size and training data volume, transformer-based language models show unprecedented capability in instruction following for text and code generation~\cite{brown2020language,sanh2021multitask,chen2021evaluating}. In the context of tool manipulation, we cast goal descriptions and optional information as an instruction in the context and task the LLMs to generate code for API calls as the continuation. 

\begin{wrapfigure}[9]{r}{0.475\textwidth}
\vspace{-23pt}
    \begin{minipage}{0.475\textwidth}
      \begin{algorithm}[H]
        \caption{API Call Generation}
        \label{alg:multi_step}
        \begin{algorithmic}[1]
            \small
            \Require{Goal $g$, API docs $\mathcal{D}$, action generator $
            \mathcal{A}$}
            \Require{Optional info $O$}
        \Procedure{ActionGen}{$g$, $\mathcal{D}$, $\mathcal{A}$, $O$}
        \State{$\mathcal{D}_g \leftarrow \mathcal{R}\left(g, 
\mathcal{D}\right) $} \Comment{Retrieve API functions}
           \State {$C_g\leftarrow\mathcal{A}\left(g, \mathcal{D}_g, O \right)$} \Comment{API call generation}
        \State\Return {$C_g$}
        \EndProcedure
        \end{algorithmic}
      \end{algorithm}
    \end{minipage}
  \end{wrapfigure}
\paragraph{Action generator} A key implementation for tool manipulation is an action generator $\mathcal{A}$
which maps a goal $g$ to API calls $ C_g$. As open-source LLMs likely have not seen the information regarding the relevant APIs, we augment an LLM $\mathcal{M}$ into an action generator by providing access to a pool of $m$ candidate API functions $\mathcal{D} =\left\{d_0, d_1, \cdots, d_m \right\}$.
Due to the input sequence length limit of LLMs, we provide an optional retriever $\mathcal{R}$ to retain a relevant subset of API documents $\mathcal{D}_g = \mathcal{R}\left(g, 
\mathcal{D}\right) \in \mathcal{D}$. 
Thus, the action generator produces the sequence of API calls $C_g = \mathcal{A}\left(g, \mathcal{D}_g, O\right)$, where $O$ represents the optional information that can be included in the prompt. 
This is a naive way of retrieval augmented generation \cite{ram2023context, lewis2020retrieval, izacard2022few} and we employ an off-the-shelf retriever implementation~\cite{bm25} for our study, but we also highly encourage the community to explore algorithms tailored for the action generator.

\paragraph{Single and multi-step tool manipulation}
As shown in \Cref{fig:task_setup}, an action generator may interact with software in either a single-step or a multi-step scenario. 
In a single-step scenario, action generator directly produces an API call sequence $C_g=\mathcal{A}\left(g, \mathcal{D}_g, \emptyset\right)$. 
In a multi-step scenario, the action generator produces a series of API call sequences $C_g = \cup_i C_{g, i}$ where each segment $C_{g, i}$ is used to interact with a predefined environment $\mathcal{E}$ and generates the observation $O_i = \mathcal{E}(C_{g, i})$. The observation is then used to generate a new segment $C_{g, i+1} = \mathcal{A}\left(g, \mathcal{D}_g, O_i\right)$. The process stops at an exit state. 
Throughout the remainder of this paper, we use the single-step setup for illustration clarity unless stated otherwise. Our experiments in~\Cref{sec:experiments} cover both single and multi-step cases.

\subsection{Motivating Observation}
\begin{wraptable}[8]{r}{0.5\linewidth}
\caption{Huge capability gaps on a house searching task. Open-source LLMs lag behind the OpenAI GPT-4 by $70\%$ on success rate.}\label{tab:motivating_observation}
\begin{adjustbox}{max width=\linewidth}
\small
\setlength{\tabcolsep}{3pt}
\begin{tabular}{ccccc}
\toprule
Model & GPT-4 & LLaMA & StarCoder & CodeGen \\
\midrule
Open source & \textcolor{red}{\xmark} & \textcolor{green}{\cmark} & \textcolor{green}{\cmark} & \textcolor{green}{\cmark} \\
Success rate & 77\% & 0\% & 7\% & 0\% \\
\bottomrule
\end{tabular}
\end{adjustbox}
\end{wraptable} 
To assess the tool manipulation capability of open-source LLMs, we compare them to OpenAI GPT-4 API using the setup discussed in~\Cref{subsec:tool_setup}. In this preliminary comparison, we initially anticipate the closed LLMs go exhibit an advantage in tool manipulation, as observed in traditional NLP tasks~\cite{liang2022holistic}. However we observe a significantly larger gap than expected. For instance, in a home search task, open-source LLMs have a hard time to generate correct API calls, resulting in a $70\%$ success rate gap compared to the zero-shot GPT-4 APIs as shown in~\Cref{tab:motivating_observation}. Such gap motivates us to study what impedes open-source LLM' performance.

\section{Challenges for open-source LLMs}
\label{sec:insights}
\begin{table}
\begin{minipage}[t]{0.45\linewidth}
\setlength{\tabcolsep}{4pt}
\scriptsize
\caption{Example of tool manipulation errors. Errors are highlighted in red.}\label{tab:typical_errors}
\begin{tabular}{cl}
\toprule
\textbf{Goal}           & \begin{tabular}[c]{@{}l@{}}\# To move the robot to position (x, y)\\ robot.move\_to(x, y)\\ \# To raise the arm by a given height\\ robot.raise\_arm(height)\\ Task: how to move a robot to (20, 30)?\end{tabular} \\
\cmidrule(lr){1-2}
\textbf{Expected results} & robot.move\_to(20, 30)                                                                                                                                                                                                                   \\
\midrule
\textbf{Wrong API}        & robot.{\color{red}\textbf{raise\_arm}}(20)                                                                                                                                                                                                                     \\
\cmidrule(lr){1-2}
\textbf{Wrong Arguments}      & robot.move\_to({\color{red}\textbf{30, 20}})                                                                                                                                                                                                                  \\
\cmidrule(lr){1-2}
\textbf{Non-executable}        & \begin{tabular}[c]{@{}l@{}} {\color{red}\textbf{You can create a robot with}} \\ {\color{red}\textbf{robot = Robot()}}\\ {\color{red}\textbf{and move it to the target location by}} \\ robot.move\_to(20, 30)\end{tabular}                                                  \\
\bottomrule
\end{tabular}
\end{minipage}
\hspace{1em}
\begin{minipage}[t]{0.525\linewidth}
\setlength{\tabcolsep}{-1pt}


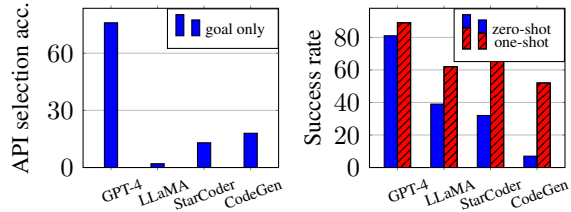
\captionof{figure}{Without API documentation exposure during inference, closed LLMs attain high accuracy in selecting APIs (left), implying potential example usage exposure during training. Hand-picked oracle one-shot demonstration improves success rate over zero-shot on the OpenWeather (right), showing the roofline impact of in-context demonstrations.}\label{fig:obsv_123}
\vspace{-3pt}
\begin{tabular}{c c}
\begin{tikzpicture}[]
    \begin{axis}[
        ybar=0.0cm,
        width=0.57\linewidth,
        height=3.7cm,
        bar width=5pt,
        enlarge x limits=0.2,
        ymajorgrids,
        ylabel={API selection acc.},
        ymin=0,
        xtick=data,
        ytick={0, 30, 60, 90},
        x tick label style={xshift=3ex,rotate=20,anchor=east,yshift=-4pt,font=\tiny},
        y label style={yshift=-2.5ex, font=\small},
        symbolic x coords = {GPT-4, LLaMA, StarCoder, CodeGen},
        tickwidth         = 1pt,
        legend style={
            at={(0.715,1)},
            anchor=north,
            legend columns=1,
            font=\tiny,
            /tikz/every even column/.append style={column sep=0.cm, row sep=-2pt},
        },
        ]
        
        \addplot[black, fill=blue] coordinates { 
            (GPT-4, 76)
            (LLaMA, 2)
            (CodeGen, 18)
            (StarCoder, 13)
        };

                    
        \legend{}
        \addlegendimage{black, fill=blue}
        \addlegendentry{goal only}
    \end{axis}
\end{tikzpicture} &
\begin{tikzpicture}[]
    \begin{axis}[
        ybar=0.0cm,
        enlarge x limits=0.2,
        width=0.57\linewidth,
        height=3.7cm,
        bar width=5,
        ymajorgrids,
        ylabel={Success rate},
        ymin=0,
        xtick=data,
        ytick={0, 20, 40, 60, 80},
        x tick label style={xshift=3ex,rotate=20,anchor=east,yshift=-4pt,font=\tiny},
        y label style={yshift=-3ex, font=\small},
        symbolic x coords = {GPT-4, LLaMA, StarCoder, CodeGen},
        tickwidth         = 1pt,
        legend style={
            at={(0.715,1)},
            anchor=north,
            legend columns=1,
            font=\tiny,
            /tikz/every even column/.append style={column sep=0.cm, row sep=-6pt},
        },
        ]
        
        \addplot[black, fill=blue] coordinates { 
            (GPT-4, 81)
            (LLaMA, 39)
            (CodeGen, 7)
            (StarCoder, 32)
        };

        \addplot[black, fill=red, postaction={pattern=north east lines}] coordinates {
            (GPT-4, 89)
            (LLaMA, 62) 
            (CodeGen, 52)
            (StarCoder, 67)
        };
                    
        \legend{,}
        \addlegendimage{black, fill=blue}
        \addlegendentry{zero-shot}
        \addlegendimage{black, fill=red, postaction={pattern=north east lines}}
        \addlegendentry{one-shot}
    \end{axis}
\end{tikzpicture}
\end{tabular}
\end{minipage}

\end{table}

\begin{wraptable}[9]{r}{0.55\textwidth}
 \vspace{-1.1em}
\captionof{table}{Categorized typical tool manipulation error types on a weather query tool.}
\label{tab:error_breakdown}
\small
 \vspace{-0.45em}
\begin{tabular}{@{}cc@{\hskip 0.5em}c@{\hskip 0.5em}c@{\hskip 0.5em}c}
\toprule
\multicolumn{1}{l}{} & GPT-4 & LLaMA & StarCoder & CodeGen \\
\cmidrule(lr){2-5}
Failure rate       & 19\% & 61\%  & 68\%  & 93\%            \\
\midrule
API selection      & 0\%  & 22\%  & 22\%  & 30\%             \\
Args. populating   & 14\% & 32\%  & 23\%  & 63\%             \\
Non-executable     & 5\%  & 7\%   & 23\%  & 0\%	           \\
\bottomrule
\end{tabular}
\end{wraptable}
To demystify key challenges, we study the behaviors of open-source LLMs in tool manipulation. By analyzing common mistakes in a weather query task,
we discover three challenges to attain strong tool manipulation capabilities. As shown in~\Cref{tab:typical_errors}, we observe that open-source LLMs often 
face difficulty in (1) API selection, (2) API argument population, and (3) generating legitimate and executable code
\footnote{If a failure case has multiple errors, we categorize it by the first triggered category in the following order: non-executable generation, wrong API selection, wrong argument populating}. 
These insights are described in detail in this section and inspire the techniques to alleviate the challenges in~\Cref{sec:techniques}. 


\paragraph{Difficulty in API selection}
We observe that API selection failures often involve using incorrect APIs and even hallucinating non-existent API names.
To quantitatively understand the intrinsic capability in API selection, we compare open-source LLMs to GPT-4 without providing any documentation or in-context demonstrations during inference. The results, as shown in Figure \ref{fig:obsv_123} for the weather query tool OpenWeather, reveal that GPT-4 can choose the right API without additional information beyond the goal, while open-source models struggle. Such capability disparity entails that \emph{closed LLMs potentially internalize knowledge of API usage during training}.

\paragraph{Confusion in populating arguments}
After the action generator selects the appropriate APIs, the subsequent challenge lies in parsing the goal description and populating the API arguments. At this stage, we observe that open-source models often provide wrong values for the required API arguments. 
The confusion in argument populating contributes to up to $63\%$ of the failures in open-source models, as shown in~\Cref{tab:error_breakdown}. In an attempt to mitigate this issue, we provide the LLMs with a hand-picked oracle in-context demonstration which achieves the same goal with different argument values. We show in~\Cref{fig:obsv_123} that the hand-picked oracle examples improve success rates by up to $45\%$. It is important to note that oracle examples are not intended as a solution for argument populating confusion, as they are hand-picked on a per-test-case basis. Nonetheless, these observations suggest that \emph{in-context demonstrations can substantially enhance open-source LLMs for tool manipulation}.

\paragraph{Non-executable generation}


The third common failure of open-source LLMs is non-executable generation. Such failures encompass issues such as language verbosity around API calls and adherence to natural language based guidelines, as shown in~\Cref{tab:typical_errors}. Open-source models sometimes exhibit such errors in $23\%$ of one hundred weather query cases.
These observations underscore \emph{the necessity of regulating open-source LLMs to exclusively generate code.}

\section{Boosting Open-source LLMs for Tool Manipulation}
\label{sec:techniques}
The insights from~\Cref{sec:insights} emphasize the importance of tuning with API usage examples, in-context demonstration and generation regulation in the domain of tool manipulation. In this section, \emph{we revisit three techniques from the LLM literature and adapt them to address the aforementioned challenges, using a practical amount of human supervision}. 
We first introduce model alignment with programatically curated data to internalize API usage knowledge in~\Cref{subset:model_align}. We then discuss augmenting open-source LLMs with an in-context demonstration retriever in~\Cref{subset:demo_retrieve}. Lastly, we apply a system prompt to regulate generation in~\Cref{subset:sysprompt}.
These techniques collectively serve as a strong baseline for alleviating the challenges presented in~\Cref{sec:insights} and inspiring further innovations.

\subsection{Multi-tool model alignment with programmatic data curation}
\label{subset:model_align}

Model alignment, through tuning LLMs with usage examples, plays a vital role in improving LLMs for capabilities such as instruction following and conversation~\cite{ouyang2022training, glaese2022improving, chung2022scaling}. 
In light of our insights from in~\Cref{sec:insights}, we recognize the potential of model alignment with API usage examples to improve API selection capability. To practically leverage such alignment for tool manipulation, it requires a data curation strategy without massive manual example writing. Towards this end, we prototype a method which generates usage examples from human-curated templates.

\begin{wrapfigure}[18]{r}{5.8cm}
\vspace{-12pt}
\caption{Programmatic training data generation using templates and random values}\label{fig:data_gen}
\centering
\vspace{-6pt}
\includegraphics[width=\linewidth]{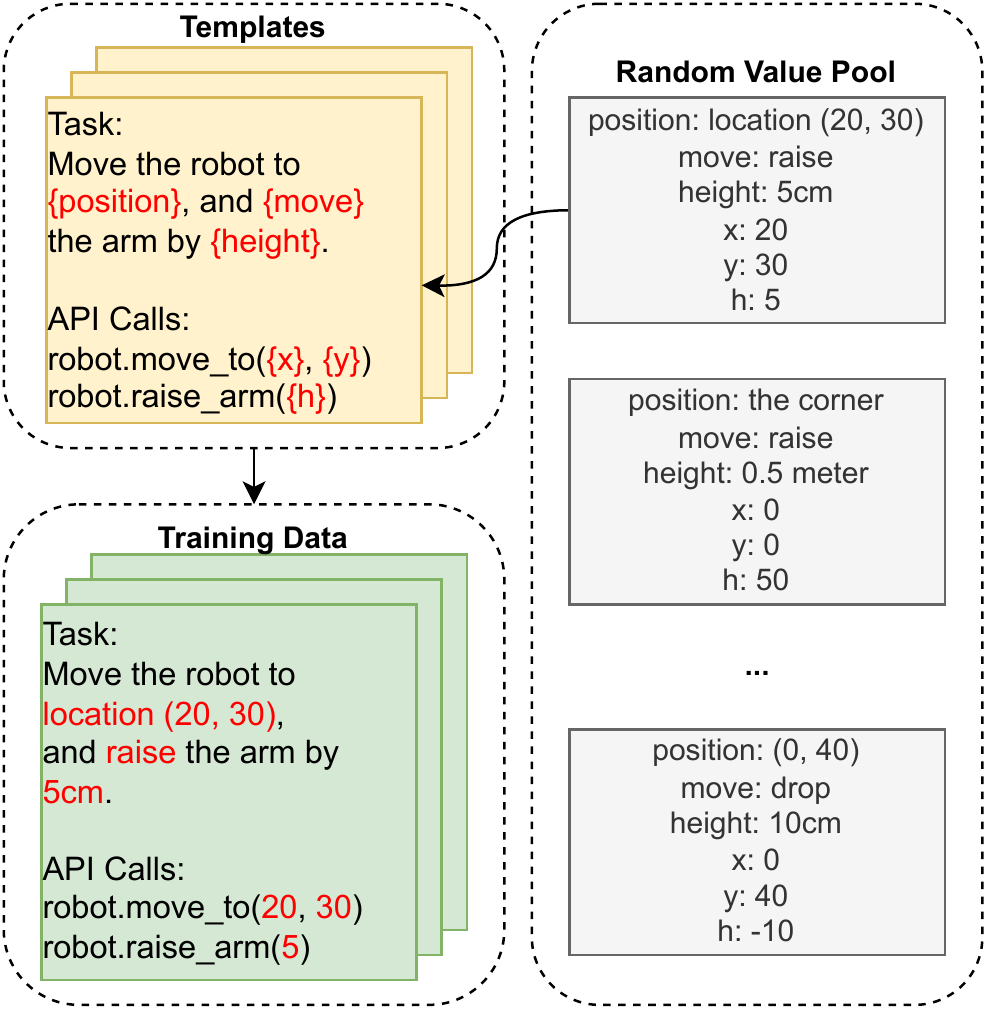}
\end{wrapfigure} 
\Cref{fig:data_gen} depicts our flow to generate alignment data.
We create a handful of templates consisting of goal descriptions and corresponding API calls. 
These templates contain one or more placeholder pairs. Each of these pairs maps to a key word in the goal and an argument in the corresponding API calls. 
We also provide a pool of candidate values for each keyword and randomly choose values to fill in the placeholders within the template. 
Given a tool with $n$ candidate APIs, we only require $\mathcal{O}(n)$ human-curated templates to ensure practical human supervision. Specifically we use a principle where each of the $n$ APIs is encouraged to appear in at least one template. 
In practice, we find it takes on average one day for one developer to curate the data for one software tool in our benchmark; this includes writing the goal templates, providing the pool of argument values and generate the data. We provide example templates we use for different tools in~\Cref{sec:app_exp_details}. 
With data curated for all the tools, we perform model alignment tuning \emph{jointly for all tools and produce a single model}.



\subsection{Demonstration retrieval}
\label{subset:demo_retrieve}

In~\Cref{sec:insights}, we demonstrate the efficacy of hand-picked oracle examples in improving argument populating. However, extending from oracles to practical in-context demonstration poses two challenges. First, given $n$ API function candidates, there are exponentially many combinations of API calls associated with different goals. Thus, LLMs should be capable of generalizing to a wide variety of goals based on a limited number of examples. Second, to ensure effective demonstration, it is important to provide LLMs with only the relevant examples without human interventions.

To fulfill the above two desiderata, we augment open-source LLMs with a demonstration retriever module. This module revolves around a repository where every API is required to appear in only one human-curated demonstration. This implies that only $\mathcal{O}(n)$ examples are needed. 
Among these demonstration examples, the retriever selects the most semantically similar examples to the goal descriptions.

\begin{wrapfigure}[13]{r}{5.5cm}
\vspace{-12pt}
\caption{In-context demonstration can improve both closed and open-source models on Home Search, a tool for browsing houses on sale.}\label{fig:demo_retrieve}
\vspace{-4pt}
\begin{tikzpicture}
    \begin{axis}[
        xlabel=$x$,
        ylabel=$y$,
        xmin=0, xmax=8,
        ymin=0, ymax=100,
        xtick={0, 1, 2, 3, 4, 5, 6, 7, 8},
        ytick={30,60,90},
        width=1.03\linewidth, height=3.5cm,
	ylabel=Success rate,
	ymajorgrids,
        y tick label style={font=\small},
        y label style={yshift=-3ex, font=\small},
	xlabel=In-context examples,
        x tick label style={font=\small},
        x label style={yshift=1.5ex, font=\small},
        legend style={
            at={(0.5,-0.45)},
            anchor=north,
            legend columns=2,
            font=\tiny,
            /tikz/every even column/.append style={column sep=0.cm},
        },
    ]
    \addplot[smooth,mark=*,blue] plot coordinates {
        (0,76.6)
        (1,89)
        (2,91)
        (3,93)
        (4,99)
        (5,97)
        (6,97)
        (7,99)
        (8,99)
    };
    \addlegendentry{GPT-4}

    \addplot[smooth,color=red,mark=square*]
        plot coordinates {
        (0,0)
        (1,45)
        (2,54)
        (3,66)
        (4,68)
        (5,67)
        (6,65)
        (7,62)
        (8,58)
    };
    \addlegendentry{LLaMA}

    \addplot[smooth,color=green,mark=triangle*]
        plot coordinates {
        (0,7)
        (1,66)
        (2,71)
        (3,82)
        (4,82)
        (5,82)
        (6,84)
        (7,86)
        (8,86)
    };
    \addlegendentry{StarCoder}


    \addplot[smooth,color=brown,mark=diamond*]
        plot coordinates {
        (0,0)
        (1,26)
        (2,40)
        (3,52)
        (4,61)
        (5,59)
        (6,52)
        (7,52)
        (8,52)
    };
    \addlegendentry{CodeGen}
    \end{axis}
\end{tikzpicture}
\end{wrapfigure}
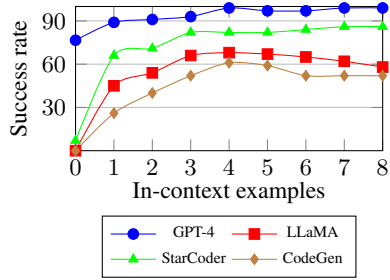 
\paragraph{Validation} 
To verify the effectiveness of demonstration examples in practice, we empirically show that the retrieved demonstrations can improve the success rate on goals requiring API combinations unseen in the example repository. 
In particular, we evaluate this approach on the home search task which exposes $15$ API functions and requires multiple functions to accomplish each goal. With only $10$ human-curated demonstrations that do not precisely match any of the $100$ test cases in terms of API combinations, the retrieved demonstrations can boost the success rate by up to $79\%$ across open-source LLMs and make GPT-4 nearly perfect, as shown in \Cref{fig:demo_retrieve}. This shows that the demonstration examples can improve tool manipulation for unseen types of goals with a repository of size $\mathcal{O}(n)$ only.



\subsection{Generation regulation with system prompts}
\label{subset:sysprompt}

\begin{wrapfigure}[17]{r}{6.2cm}
\vspace{-12pt}
\caption{System prompt with guidelines to only generate code in a desired format. Red parts are populated with real data for each test case during inference.}
\vspace{-5pt}
\label{fig:sysprompt}
\includegraphics[width=\linewidth]{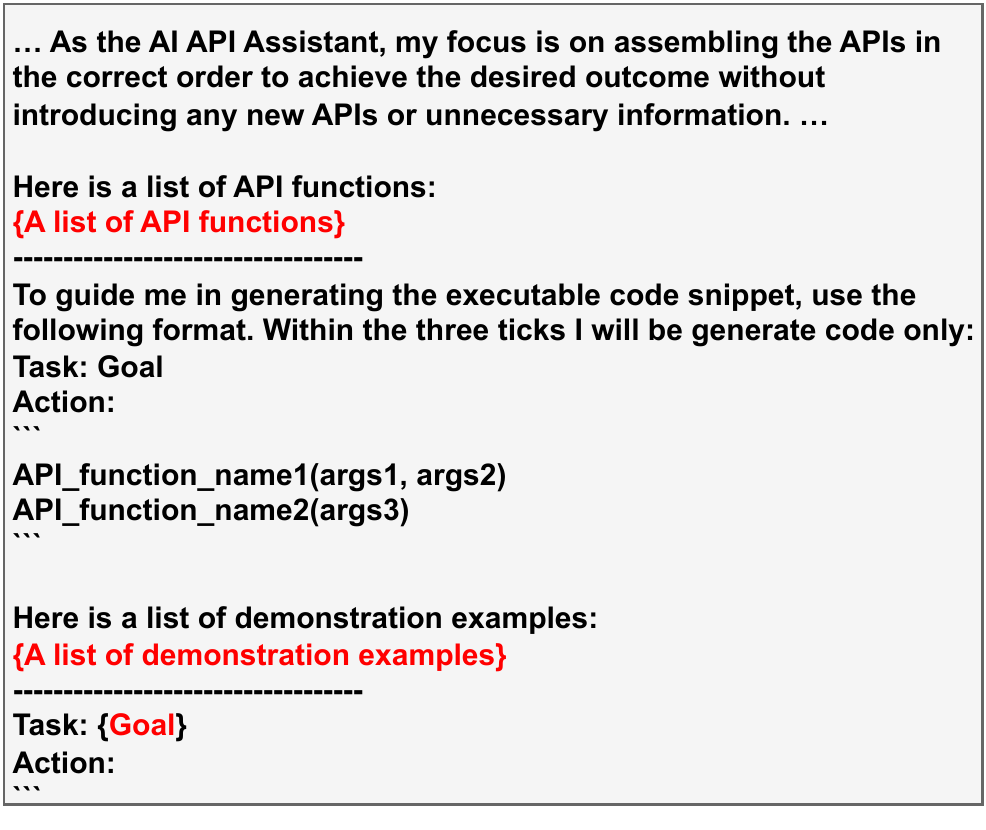}
\end{wrapfigure} 
The use of system prompts is a well-established technique in chatbots powered by LLMs~\cite{glaese2022improving}. By incorporating human-chatbot conversations, system prompts can effectively control the natural language style of the generated responses. In the context of tool manipulation, we regularize open-source LLMs to exclusively generate API calls with a system prompt in~\Cref{fig:sysprompt}, where the black part is the template shared across all tasks and the red rows are instantiated during inference for a certain goal. Our system prompt first defines a format that combines text sections containing goals, demonstrations, and generations. It then provides explicit guidelines in natural language, instructing the LLMs to generate code exclusively. The system prompt incorporates the goal description and the retrieved API functions directly for each request, reducing the human development effort to a one-time task. 

\section{\snact: A New Tool Manipulation Benchmark}
\label{sec:benchmarks}
To evaluate open-source LLMs in the domain of tool manipulation, we curate a benchmark suite from both existing datasets and newly collected ones. This benchmark stands out as the first open-source test bench with predefined test cases for quantitative evaluation, distinguishing it from recent tool manipulation research using closed LLMs \cite{li2023api, qin2023tool}. In this section, we introduce the software tools and the evaluation infrastructure. We also demonstrate the level of challenges posed by each tool, in terms of the ability to generalize to unseen API combinations and the requirement for advanced reasoning.

\subsection{Software tools and evaluation infrastructure} 
As shown in~\Cref{tab:all_tasks}, our benchmark consists of five tasks we collected and three tasks derived from existing datasets, including VirtualHome\cite{puig2018virtualhome, huang2022language}, Webshop\cite{yao2023webshop} and Tabletop\cite{liang2022code}.
They cover both single-step and multiple-step action generation, which requires selecting and combining from $2$ to $108$ API functions to accomplish the goals. 
Each task consists of approximately approximately $100$ test cases, including goal descriptions and the ground truth API calls. We also provide a limited number of demonstration examples to aid model predictions\footnote{For WebShop, we find that more than $\mathcal{O}(n)$ demonstration examples can improve the success rate. Nonetheless, these examples can be acquired from programmatic software operations without heavy human curation.}. We include a comprehensive introduction and analysis of each task within the benchmark in \Cref{sec:benchmark_details}. 

We use \emph{success rate} as the primary evaluation metric for most tasks, except for the WebShop where we report rewards, as well as for VirtualHome where we use executability and Longest Common Subsequence (LCS), following the original metrics proposed by the respective authors. To facilitate evaluation, we build an infrastructure that executes the API calls generated by the action generators and assess the final outcome. This process enables reliable evaluation of tool manipulation capabilities without restricting the action generators to perfectly match the ground truth API calls.

\subsection{Level of challenges}

\begin{table}[]
\centering
\caption{Tasks in the \snact. We provide demonstration examples for few-shot in-context-learning while test cases are for quantitatively evaluation. We develop API complexity, a metric to quantify the challenge level in generalizing to unseen API combinations; higher complexity indicates more challenging tasks. We package the challenges beyond API complexity as advanced reasoning. We refer to \Cref{sec:benchmark_details} for more details on these tasks.}
\label{tab:all_tasks}
\begin{adjustbox}{max width=\textwidth}
\begin{tabular}{@{}cccccccccc@{}}
\toprule
\multicolumn{1}{l}{}            & \multicolumn{6}{c}{\textbf{Single Step}}                                         & \multicolumn{2}{c}{\textbf{Multi-Step}}                                                             \\
\cmidrule(lr){2-7} \cmidrule(lr){8-9} 
                                & \textbf{Open}                           & \textbf{The Cat}                    & \textbf{Home}        & \textbf{Trip}        & \textbf{Google}      &                                        &                   \textbf{WebShop}                 &                                     \\
\multirow{-2}{*}{\textbf{Task}} & {\color[HTML]{1F1F1F} \textbf{Weather}} & {\color[HTML]{1F1F1F} \textbf{API}} & \textbf{Search}      & \textbf{Booking}     & \textbf{Sheets}      & \multirow{-2}{*}{\textbf{VirtualHome}} & \textbf{Long / Short} & \multirow{-2}{*}{\textbf{Tabletop}} \\
\midrule
\textbf{\textit{Data}}               &                                         &                                     &                      &                      &                      &                                        &                                    & \multicolumn{1}{l}{}                \\
API functions                   & 9                                       & 6                                   & 15                   & 20                   & 108                  & 40                                     & 2                                 & 32                \\
Demonstration examples               & 18                                      & 12                                  & 10                   & 11                   & 10                   & 83                                & 1533 / 200                        & 74                \\
Test cases                   & 100                                     & 100                                 & 100                  & 120                  & 70                   & 100                                    & 100                                & 105                                 \\
\midrule
\textbf{\emph{Level of challenges}}\\
API complexity & 2.2 & 1.4 & 7.3 & 11.1 & 8.4 & 12.3 & 0.0 & 4.6  \\
Advanced reasoning & & & &  & \checkmark & & \checkmark & \checkmark \\
\bottomrule
\end{tabular}
\end{adjustbox}
\end{table}

\begin{wraptable}[13]{r}{0.36\textwidth}
\centering
\vspace{-12pt}
\captionof{table}{A typical task of Google Sheets manipulation. It requires both selecting the correct API function and reasoning on the arguments.}
\label{tab:adv_reasoning}
\small
\begin{adjustbox}{max width=\linewidth}
\begin{tabular}{ccc}
\toprule
Product                        & Cost                           & Price                          \\ 
\cmidrule(lr){1-3}
beef                           & 1                              & 3                              \\
pork                           & 5                              & 4                              \\
chicken                        & 10                             & 11                             \\
\midrule
\multicolumn{3}{l}{Task: Update beef's price to 10.}                                             \\
\multicolumn{3}{l}{\begin{tabular}[c]{@{}l@{}}Action:\\ \textbf{worksheet.update("C2", 10)}\end{tabular}} \\ \bottomrule
\end{tabular}
\end{adjustbox}
\end{wraptable}
To assess the level of challenge, we examine  \snact\  tasks based on their API complexity and the requirement for advanced reasoning.
Intuitively, API complexity indicates the challenges in generalizing to unseen API combinations and non-default argument values. Challenges beyond API complexity then involve advanced reasoning.

\paragraph{API Complexity} 
To quantify the challenge in generalizing to unseen API combinations, we develop a task-agnostic complexity score $S\in \mathbb{R}_0^+$, where 
\begin{equation}
    S(\mathcal{T}, \mathcal{X}, \mathcal{D}) = \mathbb{E}_{t \in \mathcal{T}}\min\nolimits_{e \in \mathcal{X}} d(t, e).
\end{equation}
It averages over all the test samples in the test set $\mathcal{T}$ on the minimum distance between $t$ and any demonstration example $e$ from the example pool $\mathcal{X}$.
In particular, the distance $d(t, e)$ between each test sample $t$ and a demonstration example $e$ is negatively proportional to the probability of transforming the API combination of $e$ to match that of $t$, by randomly dropping the API functions irrelevant to $t$ and inserting the uncovered API functions required by $t$ from the API pool $\mathcal{D}$.
We refer to the details of the complexity score to \Cref{sec:app_complexity_metrics} and list their values in \Cref{tab:all_tasks}.
The score is non-negative and the higher the score is, the more complex a task is. 
Despite the fact that this complexity score reflects the challenge level of API selection, it does not capture all the difficulties of a task. A task with low complexity score can still be very challenging as it might require advanced reasoning. 
For instance, even though Webshop is challenging, the API selection complexity of it is zero. This is because there are only two API functions requiring only one argument each in Webshop, and they are both covered by the examples, so there is no API selection complexity.

\paragraph{Advanced reasoning} 
Within our benchmark, advanced reasoning encompasses challenges beyond generalizing to unseen API combinations. These challenges include non API-based coding for tasks such as Google Sheets and Tabletop, as well as decision-making based on observations returned from the WebShop environment. 
For instance, in the Google Sheets example shown in \Cref{tab:adv_reasoning}, the coordinate of the beef price's cell ("C2") cannot be easily derived from either the goal or the table itself. The action generator needs to understand the content or write additional python code to derive this coordinate before calling the API function.  
In the similar scenario, WebShop task requires the action generator to extract the exact button ID to click on the webpage given the description. 
These challenges, categorized as advanced reasoning, complement the API complexity category.



\section{Experiment}
\label{sec:experiments}
In this section, we leverage the \snact\  to empirically validate the techniques introduced in \Cref{sec:techniques}. First, to concretize the capability gap between open-source and closed LLMs, we demonstrate that OpenAI GPT-4 API can have substantially higher success rate than representative open-source LLMs in~\Cref{subsec:cap_gap}. 
We then show in~\Cref{subsec:boost} that the simple techniques in~\Cref{sec:techniques} can boost open-source LLMs to achieve success rates competitive to in-context-learning with GPT-4 APIs\footnote{GPT-4 tuning APIs were not released by the time this work is done.} in four out of the eight tasks.
Through ablation studies in~\Cref{subsec:abl}, we additionally show that model alignment does the heavy lifting for boosting open-source LLMs, while system prompt and in-context learning robustify LLMs for further improvement. 

\subsection{Experiment Setup}
To establish strong baselines, we use GPT-4 API as the representative closed LLM in our study because it attains the leading accuracy in mainstream NLP tasks. 
In our study, we compare LLAMA-30B \cite{touvron2023llama}, StarCoder \cite{li2023starcoder} and CodeGen-16B-mono \cite{nijkamp2022codegen} to GPT-4. LLAMA represents open research models, while StarCoder and CodeGen are publicly available for both research and commercial purposes. We choose these three models due to their superior performance on \snact\  among open-source models as shown in~\Cref{tab:baselines_over_models}\footnote{Surprisingly, we observe that for tool manipulations, open-source LLMs instruction-tuned for conventional NLP tasks do not outperform their base models before tuning.}. 
In our experiments, we consider the zero-shot setting as the out-of-the-box configuration where only API documentation is provided without any demonstration examples. We use this configuration to understand the initial gap in capabilities among models. We then incorporate all available techniques on top of this initial configuration to assess their benefits. 
For the original Tabletop dataset~\cite{liang2022code}, which includes examples in a few-shot setting without explicit API definitions, we only evaluate settings with in-context demonstrations. 
More detailed setup information is included in \Cref{sec:app_exp_details}. We run each job 3 times with different random seeds and report average accuracy. The variation is minimal, so we ignore them in the main paper but report them in appendix.

\subsection{Capability Gap}
\label{subsec:cap_gap}
\Cref{tab:baselines} exhibits significant disparities in tool manipulation between the closed GPT-4 API and open-source models in the out-of-the-box zero-shot setting. 
For simpler tasks, namely Open Weather and the Cat API, which require only one API call for each goal, the open-source models exhibit success rates up to $74\%$ lower than GPT-4. 
Furthermore, on all the remaining tasks other than the Webshop, none of the LLAMA, the StarCoder and the CodeGen model can reach meaningful accuracy or compare with GPT-4. 
These results highlight an opportunity to enhance open-source LLMs.

\begin{table}[]
\caption{Capability gap in tool manipulation is substantial between closed API and open-source LLMs in the out-of-the-box zero-shot setting. Using model alignment, the in-context demonstration retriever and the system prompt, open-soured LLMs attain significant boost in success rate. GPT-4 is enhanced with the retriever and system prompt.
Tabletop is only evaluated in the few-shot fashion. 
}
\label{tab:baselines}
\begin{adjustbox}{max width=\textwidth}

\setlength{\tabcolsep}{4pt}
\begin{tabular}{@{}cccccccccc@{}}
\toprule
                                          & \textbf{Open}                           & \textbf{The Cat}                    & \textbf{Home}        & \textbf{Trip}        & \textbf{Google}      &                                        & \multicolumn{2}{c}{\textbf{WebShop}}        &                                     \\
\multirow{-2}{*}{\textbf{Task}}           & {\color[HTML]{1F1F1F} \textbf{Weather}} & {\color[HTML]{1F1F1F} \textbf{API}} & \textbf{Search}      & \textbf{Booking}     & \textbf{Sheets}      & \multirow{-2}{*}{\textbf{VirtualHome}} & \textbf{Long}        & \textbf{Short}       & \multirow{-2}{*}{\textbf{Tabletop}} \\
\midrule
\textit{\textbf{Zero-shot Baseline}}      & \multicolumn{1}{l}{}                    & \multicolumn{1}{l}{}                & \multicolumn{1}{l}{} & \multicolumn{1}{l}{} & \multicolumn{1}{l}{} & \multicolumn{1}{l}{}                   & \multicolumn{1}{l}{} & \multicolumn{1}{l}{} & \multicolumn{1}{l}{}                \\
GPT-4                                          & 81.3            & 97.4            & 76.6            & 91.5            & 5.7             & 40.8 / 8.0  & \multicolumn{2}{c}{0.0}                    & -                    \\
LLaMA-30b                                     & 39.0            & 49.0            & 0.0             & 0.0            & 0.0             & 78.0 / 0.3  & \multicolumn{2}{c}{0.0}                    & -                    \\
StarCoder                                     & 32.0            & 71.0            & 7.0             & 13.3            & 5.9             & 22.0 / 3.7  & \multicolumn{2}{c}{0.0}                    & -                    \\
CodeGen-16B-mono                              & 7.0           & 78.0            & 0.0             & 0.0            & 1.4             & 4.0/ 1.0  & \multicolumn{2}{c}{0.0}                    & -                    \\

\midrule
\textit{\textbf{Enhanced w/ techniques}}      & \multicolumn{1}{l}{}                    & \multicolumn{1}{l}{}                & \multicolumn{1}{l}{} & \multicolumn{1}{l}{} & \multicolumn{1}{l}{} & \multicolumn{1}{l}{}                   & \multicolumn{1}{l}{} & \multicolumn{1}{l}{} & \multicolumn{1}{l}{}                \\
GPT-4                                      & 99.0                                    & 98.0                                & 98.0                 & 99.2                 & 68.6                 & 29.0 / 21.7                            & 0.0                  & 0.0                  & 83.8                                \\
LLaMA-30b                                     & 100.0           & 94.0            & 87.0            & 85.8            & 2.9             & 16.0 / 24.3 & 0.0             & 0.0             & 7.5             \\
StarCoder                                     & 99.0            & 97.0            & 83.0            & 80.8            & 21.2            & 31.0 / 18.4 & 0.0             & 0.0             & 13.9            \\
CodeGen-16B-mono                              & 97.7            & 99.0            & 82.0            & 77.5            & 19.8            & 29.0 / 17.2 & 0.0             & 3.5             & 16.2                    \\
\bottomrule
\end{tabular}
\end{adjustbox}
\end{table}

\subsection{Boosting open-source LLMs}
\label{subsec:boost}
To boost the open-source LLMs, we first perform model alignment using programmatially generated data. We then apply a system prompt and a 3-shot demonstration retriever during inference. Given GPT-4 does not provide tuning APIs, we enhance the out-of-the-box GPT-4 with the same system prompt and demonstration retriever as the baseline. The improvements from the combined enhancement techniques are shown in~\Cref{tab:baselines}, where the success rates of the open-source LLMs can improve up to $90\%$. As a result, the open-source models achieve competitive or better success rates on 4 out of 8 tasks, including Open Weather, the Cat API, VirturalHome and WebShop. Moreover, on Home Search and Trip Booking, the gap between the LLAMA model and the GPT-4 API is reduced to $11\%$ and $13.4\%$ respectively, compared to the initial gap of up to $91\%$. Despite the fact that open-source models are still lagging behind on the Google Sheets and Tabletop, these observations show that \emph{our recipe can significantly improve the performance of open-source LLMs and attain success rates comparable to GPT-4 API on many of the \snact\  tasks}.

\paragraph{Human supervision} To identify the practicality of an enhancement recipe, the amount of required human supervision is a crucial factor. In our approach, human supervision is primarily in the form of in-context demonstration examples and alignment data templates.
Regarding the demonstration examples, we provide $10$ to $83$ examples for each task as shown in~\Cref{tab:all_tasks}, except for WebShop given its difficulty in advanced reasoning.
As shown in~\Cref{tab:training_data}, the number of templates for alignment data is typically less than $100$ for each task. We observe that providing these supervisions takes one developer day on average, making it practical in terms of the time cost on human supervision.

\paragraph{Remaining challenges} In our experiments, we observe that the boosted open-source LLMs still have relatively low success rates on tasks that require advanced reasoning, such as Google Sheets, WebShop and Tabletop tasks. This implies the need to further enhance the reasoning capabilities of open-source models. We are excited about the prospect of more exploration from the community to address the challenges for tool manipulation on these complex tasks.


\subsection{Ablation Study}
\label{subsec:abl}

\begin{wraptable}[14]{r}{6cm}
\vspace{-12pt}
\caption{The number of \snact~tasks improved (+N) or hurt (-N) over the baselines when adding or dropping techniques.}
\label{tab:breakdown}
\begin{adjustbox}{max width=\linewidth}
\setlength{\tabcolsep}{2pt}
\begin{tabular}{@{}lccc@{}}
\toprule
               & LLaMA & StarCoder & CodeGen \\
\midrule
\textbf{Zero-shot} & - & - & -                                                \\
+ Sys. Prompt       & +4 & +4 & +4                                               \\
+ 3-shot & +8 & +8 & +8                                                \\
+ Alignment   & +7 & +7 &+7                                     \\  
\midrule
\textbf{Full system} & - & - & -                                                \\
- Sys. Prompt       & -0 & -2 & -3                                                \\
- 3-shot    & -3 & -4 & -5                                                 \\
- Alignment  & -5 & -5 & -7                                      \\        
\bottomrule
\end{tabular}
\end{adjustbox}
\end{wraptable} 



We break down the contribution of the techniques in two ways.
First, we apply each technique individually on top of the out-of-the-box zero-shot configuration and evaluate its impact. 
As shown in~\Cref{tab:breakdown}, both the 3-shot in-context demonstration and model alignment techniques bump up the success rates across all tasks, while the system prompt only benefits simple tasks that involve relatively fewer API calls for each goal.

Next, we consider the combination of all techniques and remove them one at a time to evaluate their relative contributions within the full system. As shown in in~\Cref{tab:breakdown}, solely removing model alignment triggers success rate degradation in up to 7 tasks, while removing either in-context demonstration up to 5 tasks and dropping system prompt up to 3. We notice that the tasks that are not significantly impacted when removing techniques are typically the ones with relatively low success rate (usually <20\% even in the full system). Thus, those accuracy changes are hypothetically subject to high variance and fluctuation.
The full results from the experiments in this section can be found in \Cref{tab:baselines_over_techniques}.

\section{Related work}
\label{sec:related}

Our work establishes a strong connection to the LLM-driven program synthesis. In contrast to the conventional rule-based code generation in popular compilation frameworks \cite{lattner2020mlir}, recent auto-regressive LLMs such as CodeGen\cite{nijkamp2022codegen}, SantaCoder\cite{allal2023santacoder} and StarCoder\cite{li2023starcoder} treat the problem as a sequence generation task and demonstrate superior capabilities in emitting semantically correct computer programs. We use CodeGen as a representative from these models in our study for API call generation. 

Tool manipulation are also known as tool augmented learning \cite{qin2023tool, yang2023foundation}. Some of the works seek to augment generations with the execution results from various tools\cite{schick2023toolformer, mialon2023augmented, yao2022react, izacard2022few, liang2023taskmatrix, cobbe2021training, parisi2022talm}, while another line of works focus on executing the tools themselves, including embodied robotic learning \cite{liang2022code, huang2022language, ahn2022can, singh2022progprompt, vemprala2023chatgpt}, and automation for other tools \cite{yao2023webshop, nakano2021webgpt, wu2023visual, kim2023language}. 
We focus on the study of the second stream with different models and techniques. 

Recent works in tool manipulation with LLMs mostly study techniques to enhance in-context-learning with closed LLMs APIs~\cite{schick2023toolformer, li2023api, qin2023tool, autogpt, shen2023hugginggpt}. In contrast, we study simple techniques to allow for developers to practically build on top of open-source LLMs.
The three techniques we mention in this paper~\cite{glaese2022improving, brown2020language, izacard2022few, ratner2017snorkel} are well studied in the conventional NLP tasks. We revisit and adapt them in the context of tool manipulation on open-source models with a practical amount of human supervision.
In the recent LLM literature, there are several works presenting tool manipulation benchmarks~\cite{li2023api, qin2023tool}. Compared to these benchmarks, the \snact~is the first one providing predefined test cases for evaluation on real execution results.

\section{Conclusion}
\label{sec:conclusion}

In this paper, we answer the question \emph{can we enhance open-source LLMs to compete with leading closed LLM APIs in tool manipulation, with practical amount of human supervision}. 
Drawing from our observations of the common tool manipulation failures and insights from the literature on conventional NLP tasks with LLM, we propose to instantiate model alignment with programmatical data generation, system prompts, and in-context demonstration retrievers to improve the tool manipulation capability of open-source models.
To comprehensively evaluate the impact of these techniques, we create the \textit{\snact}, a benchmark consisting of diverse software tools for real-world tasks. 
Our results demonstrate that these techniques can make the leading open-source LLMs competitive with the OpenAI GPT-4 in $4$ out of $8$ \snact\  tasks, all achieved with a practical amount of human labeling effort.

\begin{ack}
We sincerely appreciate the helpful discussion with Urmish Thakker, Tian Zhao, Raghu Prabhakar, Kaizhao Liang, Petro Junior Milan, Bowen Yang, Qinghua Li and Yaqi Zhang. 
\end{ack}

\bibliographystyle{IEEEtran}
\bibliography{ref}

\newpage
\appendix

In the appendix section, we provide detailed information on the following aspects of our study. In ~\Cref{sec:benchmark_details}, we present the background and curation details for the 8 tasks included in \snact. 
\Cref{sec:app_baselines} focuses on the performance evaluation of an extensive suite of LLMs on \snact. 
In \Cref{sec:app_exp_details}, we delve into the details of model alignment, including the process of generating the training data and training details. We also provided the full spectrum of results for the experiments in \Cref{sec:experiments}.
Finally, in \Cref{sec:app_complexity_metrics}, we introduce the API selection complexity score system, and demonstrate its effectiveness and implication in measuring task complexity.

\section{Benchmark Details}
\label{sec:benchmark_details}

\subsection{OpenWeather}

This task involves using the REST API to interact with OpenWeather website\footnote{https://openweathermap.org/api}. We include 9 types of API calls that cater to 9 categories of queries, including but not limited to retrieving current weather data in a city, obtaining air quality data at a specific longitude and latitude, and acquiring weather forecast data for a location specified by a zip code. Making each type of API calls involves correctly filling 2 to 3 required parameters (such as \texttt{lon} for longitude and \texttt{lat} for latitude) and 0 to 3 optional parameters (such as \texttt{lang} for language and \texttt{units} for units of measurement), depending on the requirements specified in each query. In total, we develop 100 unique queries for the 9 categories and 2 demonstration examples for each category. To assess the quality of the LLM's generation, we look for the first line beginning with the word "curl", if it exists. We then execute this line using the shell process. If the shell process returns a non-zero value, we declare "not executable" for this generation. On the other hand, if the code can be executed, we compare the returned response with the corresponding result from the ground-truth Curl request. The model's generation will be considered successful if the output matches the expected result precisely.

\subsection{The Cat API}

This task is a similar REST API task as the OpenWeather, but it involves making all the \texttt{GET}, \texttt{DELETE}, or \texttt{POST} request to The Cat API website\footnote{https://thecatapi.com}. There are 6 types of API calls for 6 types of queries, including deleting a cat image from the user's list of favorites, adding an image to the user's list of favorites, returning the list of favorite images, voting up or down to an image, and searching for cat images with filtering requirements. We develop 100 queries for the test set and 2 demonstration examples for each category. To evaluate the executability and success of the LLM's generation in these scenarios, we follow a similar procedure as that of the Open Weather task. It is worth noting that for queries related to removing an image from the list of favorites, we compare the LLM's generation verbatim with the ground-truth label since duplicated deletion would inevitably lead to failure if executed.

\subsection{Home Search}
This task is designed to replicate the process of searching for homes at a specific location based on certain criteria. We design the API with 15 functions, including
\begin{itemize}
  \item \texttt{set\_location} which sets the desired location;
  \item \texttt{set\_but\_or\_rent} which specifies whether the user is looking to buy or rent a home;
  \item 12 functions for setting criteria, such as home prices, number of bedrooms, and home square footage;
  \item \texttt{search} which submits the criteria to get search results.
\end{itemize}

We consider executability and f1 score of the generated action. To ensure executable searches, the agent should make a sequence of function calls that starts with \texttt{set\_location} and \texttt{set\_buy\_or\_rent}, followed by the criterion-setting functions, and then ends with a call to the \texttt{search} function. If executable, an f1 score is computed between the criteria set by the generated program and that by the ground-truth program. We develop a test set consisting of 100 queries that asked for home options with varying criteria combinations and provide 10 demonstration examples. To test the LLM's ability to utilize unseen API functions, we intentionally exclude 3 criterion-setting functions from all demonstration examples.

\subsection{Trip Booking}
The Trip Booking task is similar to the Home Search task but with more advanced dependency requirements among function calls. It simulates the process of submitting search requests for transportation tickets, hotel rooms, or both based on specific requirements like locations, dates, and the number of tickets required. We design 20 functions for the three types of booking scenarios. Depending on the scenario, some function calls may be required while others are optional. Missing any required function call or mistake the order of some function calls results in a non-executable search, while missing optional function calls lead to an unsuccessful search. We include 120 queries in the test set and provide 11 demonstration examples.

\subsection{Google Sheets}
This task is to manipulate the real worksheets from the Google Sheets\footnote{https://www.google.com/sheets/about/}, via the gspread library\footnote{https://docs.gspread.org/}. We include 100 distinct API function calls from the gspread library, but we only create tests for the most common use cases, including updating cell values, sorting, adding or deleting rows and columns, merging cells, filtering, formatting and creating pivot tables. There are 70 test cases and 10 examples in total. We also encourage the model to utilize Pandas DataFrame\footnote{https://pandas.pydata.org/docs/reference/api/pandas.DataFrame.html} and gspread-dataframe\footnote{https://gspread-dataframe.readthedocs.io/en/latest/} for advanced manipulations, by explicitly providing 8 additional API functions and certain examples for them. The manipulation is considered as correct only if both the value and the format of each cell match the expectation.

\subsection{Virtual Home}

This task is inherited from the setting of the VirtualHome\footnote{http://virtual-home.org/} simulator and asks the LLM to generate sequences of actions for completing household activities. We develop API definitions, demonstration examples, and a test set based on the list of available examples\footnote{https://github.com/huangwl18/language-planner/blob/main/src/available\_examples.json} curated in \cite{huang2022language}. The API consists of 40 functions, each of which corresponds to a specific action used in the examples. These functions can take up to two arguments, and we collect the list of valid object names for each argument based on all examples. Some examples of the functions include \texttt{Sleep()}, \texttt{Push(object)}, and \texttt{PourInto(object1, object2)}.

The original example list contains 202 household activities, represented by 5088 examples, with each example being a series of actions to complete a specific activity. However, some activities have exactly the same solution as another activity. After deduplication, we are left with 183 unique activities with non-overlapping solutions between any two activities. We randomly select 100 activities to form the test set, while the remaining 83 tasks with their 512 solutions are used as demonstration examples.

When evaluating the LLM’s generation for a given task, we consider both executability and correctness. The generation is considered executable if it can be correctly parsed into a series of valid actions, where each action involves only recognizable objects. Regarding correctness, we measure the similarity between the generated program and the ground-truth solution, using the longest common subsequence (LCS) \cite{puig2018virtualhome} normalized by the maximum length of the two. For tasks with multiple solutions, we consider the highest LCS score from any solution.

\subsection{WebShop}

This is a multi-step task inherited from Webshop \cite{yao2023webshop}, a simulated online shopping environment. 
The task requires an agent to navigate through a series of webpages to find and purchase a desired product based on a text instruction that outlines the item description.The agent can perform two primary types of actions: \texttt{search[text]}, which involves entering a text query, and \texttt{click[button]} which involves selecting a button on the page.

We generate demonstration examples based on this file\footnote{https://github.com/princeton-nlp/WebShop/blob/master/baseline\_models/data/il\_trajs\_finalized\_images.zip}, which contains trajectories collected from humans performing the online shopping tasks. We formulate each trajectory into a series of (instruction, webpage description, action) tuples in plain text format. The Long version of the demonstration set consists of 1533 full trajectories, which often exceed the input sequence length limit of the LLM. To address this issue, we provide a Short version of the demonstration examples, by first removing 80\% of the non-targeted items from any webpage description, and selecting only the 200 shortest trajectories from the complete set.

For evaluation, we use the predefined simple mode of the WebShop environment\footnote{https://github.com/princeton-nlp/WebShop\#text-environment-simple-mode} and set up the environment with the provided option of using only 1000 random products. We include 100 instructions from sessions with ID numbers 0 to 99 in the test set. We define success as making a purchase which receives a positive reward from the environment within 25 steps. 

\subsection{Tabletop}

This task is developed based on the simulated tabletop manipulation domain presented by \cite{liang2023code} and outlined in their Appendix K. In this simulation environment, a UR5e robot with a Robotiq 2F85 jaw gripper can perform pick and place actions parameterized by 2D top-down positions. We reuse their API definitions and prompts as demonstration examples. We iterate on the 14 instruction templates used in their evaluation benchmark and create 15 types of tasks that involve manipulating up to 4 colored blocks and 4 colored bowls. For each type of task, we generate 7 valid initial setups of blocks and bowls for the test set, ensuring that no collisions occur during the execution of a valid solution. The success of the LLM’s generated program is determined by whether all objects are within a small threshold of their target positions after execution.

\section{Comprehensive Model Evaluation on the \snact}
\label{sec:app_baselines}

\begin{table}[]
\centering
\caption{The achitecture and training data of all the models in our evaluation. The models are grouped by their architecture and training data.}
\label{tab:all_models}
\begin{adjustbox}{max width=\textwidth}

\setlength{\tabcolsep}{2pt}
\begin{tabular}{ccccccc}
\toprule
\multicolumn{1}{c}{\multirow{2}{*}{\textbf{Model}}} & \multicolumn{3}{c}{\textbf{Architecture}}                                                                             & \multicolumn{3}{c}{\textbf{Data}}                                                                                                                                                                                       \\
\cmidrule(lr){2-4} \cmidrule(lr){5-7}
\multicolumn{1}{c}{}                                & \multicolumn{1}{c}{\textbf{Family}} & \multicolumn{1}{c}{\textbf{Size}} & \multicolumn{1}{c}{\textbf{Max SS}} & \multicolumn{1}{c}{\textbf{\# Tokens}} & \multicolumn{1}{c}{\textbf{Pretraining}}                                                                                  & \multicolumn{1}{c}{\textbf{Finetuning}} \\
\midrule
\multicolumn{1}{c}{\textit{\textbf{Closed-source}}}          &                                     &                                   &                                     &                                        &                                                                                                                           &                                                    \\
text-davinci-003                                    & gpt3                                & 175b                              & 4096                                & -                                      & -                                                                                                                         & -                                                  \\
gpt-3.5-turbo                                       & gpt3                                & -                                 & 4096                                & -                                      & -                                                                                                                         & -                                                  \\
text-curie-001                                      & gpt3                                & 6.7b                              & 2048                                & -                                      & -                                                                                                                         & -                                                  \\
gpt4                                                & gpt4                                & -                                 & 8192                                & -                                      & -                                                                                                                         & -                                                  \\
\midrule
\multicolumn{1}{c}{\textit{\textbf{Open-source}}}           &                                     &                                   &                                     &                                        &                                                                                                                           &                                                    \\
bloomz                                              & bloom                               & 176b                              & 2048                                & 366B                                   & bloom corpus                                                                                            & xP3                                                \\
\cmidrule(lr){1-7}
llama-65b                                           & llama                               & 65b                               & 2048                                & 1.4T                                   & \multirow{4}{*}{\begin{tabular}[c]{@{}c@{}}CCNet, C4, \\GitHub, Wikipedia, \\ Books, ArXiv, \\ Stack Exchange\end{tabular}} & -                                                  \\
llama-30b                                           & llama                               & 30b                               & 2048                                & 1.4T                                   &                                                                                                                           & -                                                  \\
llama-13b                                           & llama                               & 13b                               & 2048                                & 1.4T                                   &                                                                                                                           & -                                                  \\
llama-13b-alpaca                              & llama                               & 13b                               & 2048                                & 1.4T                                   &                                                                                                                           & GPT-4 responses, Alpaca                                                  \\
\cmidrule(lr){1-7}
starcoderbase                                        & bigcode                                & 15.5b                               & 8192                                &             1T                       &      \multirow{2}{*}{The Stack}                                                                                                                     &    -  \\
starcoder                                        & bigcode                                & 15.5b                               & 8192                                &             1T                       &                                                                                                                          &    The Stack (Python)  \\
\cmidrule(lr){1-7}
opt-30b                                         & opt                                 & 30b                               & 2048                                & 300B                                   & \multirow{4}{*}{\begin{tabular}[c]{@{}c@{}}The Pile, BookCorpus, \\ CC-Stories, Reddit, \\ CCNewsV2\end{tabular}}         &             -         \\
opt-1.3b                                        & opt                                 & 1.3b                              & 2048                                & 300B                                   &                                                                                                                           &                                   -                 \\
opt-iml-30b                                             & opt                                 & 30b                               & 2048                                & 300B                                   &                                                                                                                           &                                  \multirow{2}{*}{OPT-IML Bench}                  \\
opt-iml-1.3b                                            & opt                                 & 1.3b                              & 2048                                & 300B                                   &                                                                                                                           &                                                    \\
\cmidrule(lr){1-7}
gpt-neox-20b                                            & neox                                & 20b                               & 2048                                & 450B                                   & \multirow{8}{*}{The Pile}                                                                                                 & -                                                  \\
GPT-NeoXT-Chat-Base-20B                             & neox                                & 20b                               & 2048                                &   460B                                     &                                                                                                                           & OpenChatKit IT          \\
codegen-16B-nl                                      & neox                                & 16b                               & 2048                                & 700B                                   &                                                                                                                           & -                                                  \\
codegen-16B-multi                                   & neox                                & 16b                               & 2048                                & 1T                                     &                                                                                                                           & BigQuery                                           \\
codegen-16B-mono                                    & neox                                & 16b                               & 2048                                & 1T                                     &                                                                                                                           & BigQuery, BigPython                               \\
pythia-12b                                          & neox                                & 12b                               & 2048                                & 300B                                   &                                                                                                                           & -                                                  \\
dolly-v2-12b                                        & neox                                & 12b                               & 2048                                & 300B                                   &                                                                                                                           & Dolly IT     \\
pythia-6.9b/2.8b1.4b                                          & neox                                & multi                               & 2048                                & 300B                                   &                                                                                                                           & -                                                  \\
\cmidrule(lr){1-7}
stablelm-base-alpha-7b                                        & neox                                & 7b                               & 4096                                &             800B                       &      \multirow{4}{*}{The Pile (1.5T)}                                                                                                                     &    -  \\
stablelm-base-alpha-3b                                        & neox                                & 3b                               & 4096                                &             800B                       &                                                                                                                          &    -  \\
stablelm-tuned-alpha-7b                                        & neox                                & 7b                               & 4096                                &             800B                       &                                                                                                                           & \multirow{2}{*}{\begin{tabular}[c]{@{}c@{}}Alpaca, GPT4All,\\Anthropic, Dolly, ShareGPT\end{tabular}}     \\
stablelm-tuned-alpha-3b                                        & neox                                & 3b                               & 4096                                &             800B                       &                                                                                                                           &      \\
\bottomrule
\end{tabular}
\end{adjustbox}
\end{table}

In this section, we want to compare the performance of different models on the \snact. Specifically, we selected 27 representative LLMs from both closed and open-source community, and evaluate them on the \snact~in 3-shot scenario.

\subsection{Models}

As listed in Table \ref{tab:all_models}, we select a set of representative LLMs from both closed-source and open-source community.

The closed models are the (Generative Pre-trained Transformer) GPT series from OpenAI, especially the GPT-3\cite{brown2020language} and its successors\cite{OpenAI2023-ov}. GPT-3 is a state-of-the-art language model developed by OpenAI, with 175 billion parameters, making it the largest and most powerful language model ever created. It is capable of performing a wide range of natural language processing tasks and has the potential to revolutionize the way we interact with and understand language. Due to the lack of detailed information about its training, we are motivated to study methods to build models achieving similar capabilities, especially using open-source models.

We select the representative and the most advanced open-source models from recent years in our work. They are all decoder-only models, based on transformers\cite{vaswani2017attention} architecture. 
Bloomz\cite{muennighoff2022crosslingual} is the largest open-source LLM built upon the large-scale multilingual pretrained BLOOM\cite{bigscience_workshop_2022}. Bloomz is funtuned on xP3\cite{muennighoff2022crosslingual}, a crosslingual task mixture, for crosslingual generalization to unseen tasks and languages. 
StarCoder\cite{li2023starcoder} is a family of models developed for purely code generation and synthesis with 8K context length. They exhibit superior performance on common code generation benchmarks. 
LLaMA\cite{touvron2023llama} is a family of pretrained models, that are performant on quite a few NLP benchamrks. Although they are not as large as Bloomz, they are all trained for almost 4 times longer than Bloom. This is an important reason why they are able to outperform several top peer models on many NLP tasks. Alpaca \cite{chavinlo-gpt4-x-alpaca, alpaca} is fine-tuned LLaMA-13b model on 52K instruction-following data as well responses from GPT-4. 
OPT-IML\cite{iyer2022opt} is the finetuned version of the original OPT\cite{zhang2022opt}, which is the first family of large-scale (176 billion parameters) open-source models that are trained on publicly available datasets. OPT-IML significatly improves the instruction following capability of OPT by training on a large benchmark of 2000 NLP tasks for Instruction MetaLearning (IML). We only select the publicly accessibile checkpoints from the OPT families in our work.

Another important family of models are all developed from the NeoX toolkit\cite{gpt-neox-library} and pretrained using the PILE dataset\cite{gao2020pile}. 
GPT-NeoX-20B\cite{black2022gpt} is only pretrained on the PILE, while GPT-NeoXT-Chat-Base-20B\cite{openchatkit} is further finetuned on the OIG-43M\cite{oig-data}, a dataset targetting better instruction following capability. 
CodeGen family\cite{nijkamp2022codegen} is designed for superior capability on code generation, as they are heavily finetuned on large code datasets. 
Pythia family\cite{biderman2023pythia} is a suite of models designed for analyzing LLMs across training and scaling. They are all pretrained on the Pile in the same way, but have different model sizes and intermediate checkpoints released during training. We use those variants in our ablation study.
Dolly\cite{dolly} is finetuned beyond Pythia-12b on a new, high-quality human generated instruction following dataset, crowdsourced among Databricks employees.
The StableLM family\cite{stablelm} is pre-trained on an experimental version of the PILE datasets which has 1.5 trillion tokens in total. The models have a sequence length of 4096 to push beyond the context window limitations of the existing open-source language models. The instruction tuned counterpart of each model is also released. By the time we publish this work, only 7b and 3b models are released, while the team behind them is training larger models. 

There are other notable models, such as FlanT5\cite{chung2022scaling}, the T0 family\cite{sanh2021multitask}, and the T5 family\cite{raffel2020exploring}, that have shown promising performance. We do not include all of them in our baseline comparison, as some of their features are not designed for the task at hand. For example, their tokenizers do not distinguish between spaces, tabs and new lines, making it hard for them to generate executable code based on API function calls.

\subsection{Evaluation}


To collect the baseline results, we exploit the naive approach described in section \ref{sec:background} as the action generator. 
We give each LLM sufficient max tokens to generate on each task and retrieve as many API functions as possible in the prompt. The detailed information is listed in \Cref{tab:baselines_over_models}. We evaluate all the models on a mixture of GPUs and RDUs\cite{koeplinger2018spatial, prabhakar2017plasticine, prabhakar2021sambanova}. In particular, the 176b-parameter bloomz is evaluated on RDU, while all the other models are evaluated on NVIDIA A100 GPUs with 80GB RAM. 

For these models, We only conduct the few-shot evaluation described \Cref{sec:experiments} because 1) zero-shot results are not representative, as most of them are zero, 2) it is not practical to tune all the models on our training data, and 3) few-shot results can be used as a great proxy of the model performance in all the other settings.
For the conversation-oriented models, including gpt-3.5-turbo, chavinlo/gpt4-x-alpaca, GPT-NeoXT-Chat-Base-20B and dolly-v2-12b, we additionally add \texttt{<human>:} and \texttt{<bot>:} key words in the prompt to better align with their training data format for better performance.

\begin{table}[]
\centering
\caption{The performance on \snact~of different models in 3-shot scenario. The models are group by their architecture and training data.}
\label{tab:baselines_over_models}
\begin{adjustbox}{max width=\textwidth}

\setlength{\tabcolsep}{3pt}
\begin{tabular}{@{}cccccccccc@{}}
\toprule
                                 & \textbf{Open}                           & \textbf{The Cat}                    & \textbf{Home}        & \textbf{Trip}        & \textbf{Google}      &                                        & \multicolumn{2}{c}{\textbf{WebShop}}                                   &                                     \\
\multirow{-2}{*}{\textbf{Task}}  & {\color[HTML]{1F1F1F} \textbf{Weather}} & {\color[HTML]{1F1F1F} \textbf{API}} & \textbf{Search}      & \textbf{Booking}     & \textbf{Sheets}      & \multirow{-2}{*}{\textbf{VirtualHome}} &  \textbf{Long} & \textbf{Short} & \multirow{-2}{*}{\textbf{Tabletop}} \\
max tokens to generate & 128 & 128 & 128 & 300 & 256 & 128 & \multicolumn{2}{c}{128} & 256 \\
num API function       & all & all & all & all & 10  & 10  & \multicolumn{2}{c}{all} & 0  \\
\midrule
\textbf{\textit{Closed-source}}                         & \multicolumn{1}{l}{}                    & \multicolumn{1}{l}{}                & \multicolumn{1}{l}{} & \multicolumn{1}{l}{} & \multicolumn{1}{l}{} & \multicolumn{1}{l}{}                   & \multicolumn{1}{l}{}       & \multicolumn{1}{l}{}      & \multicolumn{1}{l}{}                \\
gpt4                    & 93.0 & 96.0 & 97.0 & 96.7 & 62.9 & 23.0 / 23.5 & \multicolumn{2}{c}{0.0} & 81.0 \\
text-davinci-003        & 99.0 & 98.0 & 97.0 & 89.2 & 62.9 & 31.0 / 25.1 & \multicolumn{2}{c}{0.0} & 66.7 \\
gpt-3.5-turbo           & 90.0 & 92.0 & 80.0 & 85.8 & 51.4 & 20.0 / 18.9 & 0.0        & 1.8        & 33.3 \\
text-curie-001          & 8.0  & 58.0 & 6.0  & 6.7  & 1.4  & 12.0 / 4.1  & 0.0        & 0.0        & 1.0  \\
\midrule
\textbf{\textit{Open-source}}                         & \multicolumn{1}{l}{}                    & \multicolumn{1}{l}{}                & \multicolumn{1}{l}{} & \multicolumn{1}{l}{} & \multicolumn{1}{l}{} & \multicolumn{1}{l}{}                   & \multicolumn{1}{l}{}       & \multicolumn{1}{l}{}      & \multicolumn{1}{l}{}                \\
llama-65b               & 90.0 & 80.0 & 84.0 & 65.8 & 32.9 & 32.0 / 20.3 & 0.0 & 41.2 & 30.5 \\
llama-30b               & 78.0 & 84.0 & 66.0 & 45.0 & 37.1 & 27.0 / 21.7 & 0.0 & 30.6 & 34.3 \\

llama-13b               & 70.0 & 74.0 & 45.0 & 35.8 & 5.7  & 28.0 / 18.9 & 0.0 & 27.6 & 17.1 \\
llama-13b-alpaca   & 62.0 & 43.0 & 44.0 & 40.8 & 11.4 & 1.0 / 1.6   & 0.0 & 2.7  & 9.5  \\
\cmidrule(lr){1-1}
starcoder               & 91.0 & 84.0 & 82.0 & 51.7 & 48.0 & 23.0 / 19.4 & 2.6 & 0.0  & 21.9 \\
starcoderbase           & 90.0 & 86.0 & 79.0 & 63.3 & 42.9 & 24.0 / 16.3 & 5.8 & 23.1 & 17.1 \\
\cmidrule(lr){1-1}
codegen-16B-nl          & 51.0 & 75.0 & 37.0 & 21.7 & 7.1  & 43.0 / 18.0 & 0.0 & 0.0  & 16.2 \\
codegen-16B-multi       & 56.0 & 75.0 & 47.0 & 7.5  & 21.4 & 31.0 / 14.1 & 0.0 & 0.5  & 8.6  \\
codegen-16B-mono        & 63.7 & 72.0 & 52.0 & 28.3 & 31.5 & 28.0 / 15.7 & 1.5 & 6.6  & 15.2 \\
\cmidrule(lr){1-1}
bloomz                  & 58.0 & 85.0 & 36.0 & 22.5 & 14.3 & 9.0 / 4.9   & 0.0 & 1.0  & 1.0  \\
\cmidrule(lr){1-1}
opt-iml-30b             & 44.0 & 48.0 & 5.0  & 3.3  & 2.9  & 13.0 / 8.3  & 0.0 & 0.0  & 1.0  \\
opt-30b                 & 46.0 & 35.0 & 2.0  & 3.3  & 8.6  & 24.0 / 11.7 & 0.0 & 0.0  & 1.0  \\
opt-iml-1.3b            & 20.0 & 28.0 & 0.0  & 0.0  & 4.3  & 13.0 / 3.1  & 0.0 & 0.0  & 1.0  \\
opt-1.3b                & 18.0 & 30.0 & 0.0  & 0.0  & 1.4  & 31.0 / 9.7  & 0.0 & 0.0  & 1.0  \\
\cmidrule(lr){1-1}
neox-20b                & 55.0 & 69.0 & 27.0 & 10.8 & 18.6 & 28.0 / 15.3 & 0.0 & 8.8  & 6.7  \\
GPT-NeoXT-Chat-Base-20B & 43.0 & 73.0 & 28.0 & 10.8 & 4.3  & 26.0 / 13.1 & 0.0 & 0.7  & 7.6  \\
\cmidrule(lr){1-1}
pythia-12b              & 53.0 & 65.0 & 12.0 & 0.8  & 11.4 & 17.0 / 12.1 & 0.0 & 0.0  & 1.9  \\
dolly-v2-12b            & 0.0  & 1.0  & 10.0 & 5.0  & 7.1  & 11.0 / 8.9  & 0.0 & 0.0  & 7.6  \\
pythia-12b              & 53.0 & 65.0 & 12.0 & 0.8  & 11.4 & 17.0 / 12.1 & 0.0 & 0.0  & 1.9  \\
pythia-6.9b  & 41.0 & 72.0 & 8.0  & 7.5  & 4.3  & 29.0 / 14.0 & 0.0 & 0.0  & 8.6  \\
pythia-2.8b  & 49.0 & 54.0 & 7.0  & 3.3  & 12.9 & 24.0 / 14.8 & 0.0 & 0.0  & 7.6  \\
pythia-1.4b  & 37.0 & 48.0 & 4.0  & 5.0  & 10.0 & 22.0 / 10.7 & 0.0 & 5.2  & 7.6  \\
\cmidrule(lr){1-1}
stablelm-base-alpha-7b  & 22.0 & 47.0 & 0.0  & 0.0  & 4.3  & 28.0 / 10.3 & 0.0 & 0.0  & 2.9  \\
stablelm-tuned-alpha-7b & 23.0 & 38.0 & 0.0  & 0.0  & 1.4  & 26.0 / 7.3  & 0.0 & 0.0  & 3.8  \\
stablelm-base-alpha-3b  & 6.0  & 28.0 & 0.0  & 0.0  & 1.4  & 29.0 / 5.3  & 0.0 & 0.0  & 1.0  \\
stablelm-tuned-alpha-3b & 14.0 & 31.0 & 0.0  & 0.8  & 0.0  & 8.0 / 5.6   & 0.0 & 0.0  & 1.0  \\
\bottomrule
\end{tabular}
\end{adjustbox}
\end{table}
After we get the completion from the LLMs given a prompt, only minimal post-processing steps are applied to the completion: 
1) Properly truncate the completion, given the list of task-specific stop sequences and 
2) Replace the \texttt{\{API\_KEY\}} keywords in the completion with the real API key, so as to execute the code properly. Finally, as shown in Figure \ref{fig:task_setup}, to validate the action generated for the single-step tasks, we execute the generated API calls and compare its output against the ground truth; while for the multi-step tasks, the actions are used to interact with the environment directly and only the final status is evaluated. For each task, we report the metrics described in~\Cref{sec:benchmarks} for each task. Note that we only evaluate the top 1 generated action with sampling disabled. This is because, in practice, action can only be executed once and there is no chance to reset things and try another action.

\subsection{\snact~performance of different models}
The performance of different models are summarized in \Cref{tab:baselines_over_models}. Below we show several observations.

\paragraph{Capability Gap} Currently, the GPT family of models stands out as the leading players in the field, and there is a significant gap between GPT-4, GPT-3.5 and all the other open-source models. While open-source models may demonstrate competitiveness on some simpler tasks, they lag far behind on more challenging tasks such as Google Sheets and Tabletop.

\paragraph{Instruction tuning on conventional NLP tasks doesn't help}
Comparing the models between chavinlo/gpt4-x-alpaca and LLaMA-13b, OPT-IML and OPT, StableLM-tuned and StableLM-base, NeoX-Chat-Base-20b and NeoX, and dolly and pythia, the former model in each pair is intentionally optimized to enhance instruction following capability compared to the latter model. However, no significant accuracy improvement is observed on the \snact. 
Further, the LLaMA family, despite not undergoing any specific instruction tuning during training, still achieves relatively good quality compared to other public models.  

\paragraph{Model size is important} By comparing the performance of models from GPT faimily, LLaMA family, OPT family, Pythia family and StableLM family, we can clearly see the trend that the larger models tend to perform better on the \snact, given the same quantity and quality of their training data. 

\paragraph{Code generation is important} StarCoder and CodeGen faimily stand out among other models with similar sizes on \snact, while StarCoderBase is even on par with the llama-65b model which is more than 4 times larger in size. CodeGen-16B-mono is overall better than its base model CodeGen-16B-nl, which is not specifically tuned for code generation. It is also surprisingly better than CodeGen-16B-multi on almost all the tasks, indicating that it is highly beneficial for action generation if the model is heavily tuned on Python-style code generation. 






\section{Experiment Details}
\label{sec:app_exp_details}
In this section, we extended \Cref{sec:experiments} with more details about model training and results.

\subsection{Training data}\label{sec:training_data_section}

For the OpenWeather, The Cat API, Trip Booking, and Home Search tasks, we generate the training data by converting or expanding the demonstration examples of each task into templates and populating them with various sets of variable values. For the remaining four tasks, we format the training samples directly from the demonstration example set described in~\cref{sec:benchmarks}. We exclude any test samples from the training data and minimize the overlap of the API function call combinations between any training and test samples. For example, we make sure that the API function combinations used in each test case for the Home Search task are never present in the training data. However, for the OpenWeather task, it was unavoidable to have some overlap because each test case only involved a single function call and the training examples covered all the API functions. The numbers of templates and training samples for each task are summarized in~\cref{tab:training_data}. Example templates and variable values are shown in~\cref{tab:templates}. The training sets for all tasks, except for the Google Sheets and WebShop task, reduce the complexity score of their respective test sets when compared to the example sets. As expected, the model's accuracy shows improvement after fine-tuning.

\begin{table}[]
\centering
\caption{The statistics of model alignment data}
\label{tab:training_data}
\begin{adjustbox}{max width=\textwidth}
\begin{tabular}{ccccccccc}
\toprule
\multirow{2}{*}{\textbf{Task}} &
  \textbf{Open} &
  \textbf{The Cat} &
  \textbf{Home} &
  \textbf{Trip} &
  \multicolumn{1}{c}{\textbf{Google}} &
  \multicolumn{1}{c}{\multirow{2}{*}{\textbf{VirtualHome}}} &
  \multicolumn{1}{c}{\multirow{2}{*}{\textbf{WebShop}}} &
  \multicolumn{1}{c}{\multirow{2}{*}{\textbf{Tabletop}}} \\
 &
  \textbf{Weather} &
  \textbf{API} &
  \textbf{Search} &
  \textbf{Booking} &
  \multicolumn{1}{c}{\textbf{Sheets}} &
  \multicolumn{1}{c}{} &
  \multicolumn{1}{c}{} &
  \multicolumn{1}{c}{} \\
\midrule
Templates        & 90   & 40   & 100  & 30   & 1   & 1   & 2    & 1 \\
Repeat           & 20   & 45   & 18   & 60   & 118 & 512 & 900  & 74 \\
Training samples & 1800 & 1800 & 1800 & 1800 & 118 & 512 & 1800 & 74 \\
\midrule
Complexity score & 1.1 & 1.0 & 6.4 & 10.1 & 12.1 & 12.3 & 0.0 & 4.6 \\

\bottomrule
\end{tabular}
\end{adjustbox}
\end{table}
\begin{table}
\centering
\caption{Training template examples of different tools}
\label{tab:templates}
\begin{adjustbox}{max width=\textwidth}
\setlength{\tabcolsep}{1pt}
\begin{tabular}{@{}clll@{}}
\toprule
\multicolumn{1}{l}{}                                                                   & \multicolumn{1}{c}{\textbf{Goal}}                                                                                                                                                                 & \multicolumn{1}{c}{\textbf{Action}}                                                                                                                                                                                                                                                                                                                               & \multicolumn{1}{c}{\textbf{Variable values}}                                                                                                                                                                                                                                                                                                                     \\
\midrule
{\color[HTML]{1F1F1F} \textbf{\begin{tabular}[c]{@{}c@{}}Open\\ Weather\end{tabular}}} & \begin{tabular}[c]{@{}l@{}}What is the present weather situation \\ in {\color[HTML]{FF0000}\{city\}}? Please respond in {\color[HTML]{FF0000}\{lang\}} \\ and use {\color[HTML]{FF0000}\{units\}} units.\end{tabular}                                              & \begin{tabular}[c]{@{}l@{}}curl -X GET 'https://api.openweathermap.org/data/\\ 2.5/weather?q={\color[HTML]{FF0000}\{city\_formatted\}}\&appid=\\ \{API\_KEY\}\&lang={\color[HTML]{FF0000}\{lang\_abbr\}}\&units={\color[HTML]{FF0000}\{units\}}'\end{tabular}                                                                                                                                                                       & \begin{tabular}[c]{@{}l@{}}\{city: "Palo Alto", \\ city\_formatted: "palo+alto", \\ lang: "English", \\ lang\_abbr: "en", \\ units: "imperial"\}\end{tabular}                                                                                                                                                                                                    \\
\midrule
{\color[HTML]{1F1F1F} \textbf{\begin{tabular}[c]{@{}c@{}}The Cat\\ API\end{tabular}}}  & \begin{tabular}[c]{@{}l@{}}Add the cat photo with id={\color[HTML]{FF0000}\{image\_id\}} \\ to my list of favorites.\end{tabular}                                                                                       & \begin{tabular}[c]{@{}l@{}}curl -X POST 'https://api.thecatapi.com/v1/favourites' \\ --data '\{"image\_id":"{\color[HTML]{FF0000}\{image\_id\}"\}}'\end{tabular}                                                                                                                                                                                                                        & \{image\_id: "MTUyNTA1OA"\}                                                                                                                                                                                                                                                                                                                                      \\
\midrule
\textbf{\begin{tabular}[c]{@{}c@{}}Home\\ Search\end{tabular}}                         & \begin{tabular}[c]{@{}l@{}}Looking for homes for sale in\\  {\color[HTML]{FF0000}\{location\}} with {\color[HTML]{FF0000}\{num\_beds\}} \\ bedrooms and {\color[HTML]{FF0000}\{num\_baths\}} bathrooms, \\ between \${\color[HTML]{FF0000}\{min\_price\}} and \${\color[HTML]{FF0000}\{max\_price\}}.\end{tabular}   & \begin{tabular}[c]{@{}l@{}}API.set\_location({\color[HTML]{FF0000}\{location\}})\\ API.set\_buy\_or\_rent("buy")\\ API.set\_num\_beds({\color[HTML]{FF0000}\{num\_beds\}})\\ API.set\_num\_baths({\color[HTML]{FF0000}\{num\_baths\}})\\ API.set\_min\_price({\color[HTML]{FF0000}\{min\_price\}})\\ API.set\_max\_price({\color[HTML]{FF0000}\{max\_price\}})\\ API.search()\end{tabular}                                                                                      & \begin{tabular}[c]{@{}l@{}}\{location: "Palo Alto", \\ num\_beds: 4, \\ num\_baths: 5, \\ min\_price: 7000000, \\ max\_price: 8000000\}\end{tabular}                                                                                                                                                                                                             \\
\midrule
\textbf{\begin{tabular}[c]{@{}c@{}}Trip\\ Booking\end{tabular}}                        & \begin{tabular}[c]{@{}l@{}}Search for {\color[HTML]{FF0000}\{means\_of\_transportation\}} \\ tickets for {\color[HTML]{FF0000}\{num\_adults\}} adults \\ from {\color[HTML]{FF0000}\{location\_from\}} to \\ {\color[HTML]{FF0000}\{location\_to\}}, on {\color[HTML]{FF0000}\{departure\_date\}}.\end{tabular} & \begin{tabular}[c]{@{}l@{}}API.select\_booking\_type("trip tickets")\\ API.select\_transportation({\color[HTML]{FF0000}\{means\_of\_transportation\}})\\ API.set\_num\_adults({\color[HTML]{FF0000}\{num\_adults\}})\\ API.set\_origin(Loc({\color[HTML]{FF0000}\{location\_from\}}))\\ API.set\_destination(Loc({\color[HTML]{FF0000}\{location\_to\}}))\\ date = Date({\color[HTML]{FF0000}\{departure\_date\}})\\ API.set\_departure\_date(date)\\ API.search()\end{tabular} & \begin{tabular}[c]{@{}l@{}}\{means\_of\_transportation: "flight", \\ max\_price\_ticket: 150, \\ num\_adults: 2, \\ location\_from: "San Francisco", \\ location\_to: "Los Angeles", \\ departure\_date: "2023-08-15"\}\end{tabular}                                                                                                                             \\
\midrule
\textbf{\begin{tabular}[c]{@{}c@{}}Google\\ Sheet\end{tabular}}                        & {\color[HTML]{FF0000} \{task\}}                                                                                                                                                                   & {\color[HTML]{FF0000} \{action\}}                                                                                                                                                                                                                                                                                                                                 & \begin{tabular}[c]{@{}l@{}}\{task: "\\ | Product | Cost | Price |\\ | beef | 1 | 3 |\\ | pork | 5 | 4 |\\ | chicken | 10 | 11 |\\ | lamb | 3 | 15 |\\ | duck | 12 | 2 |\\ | fish | 2 | 100 |\\ \\ Task: \\ Sum B1:B4 and write the result below B4\\ Action:", \\ action: "\\ worksheet.update('B5', '=SUM(B1:B4)', \\ raw=False)"\}\end{tabular} \\
\midrule
\textbf{VirtualHome}                                                                   & {\color[HTML]{FF0000} \{task\}}                                                                                                                                                                   & {\color[HTML]{FF0000} \{action\}}                                                                                                                                                                                                                                                                                                                                 & \begin{tabular}[c]{@{}l@{}}\{task: "\\ Task: Read book\\ Action:", \\ action: "\\ Agent.Find(novel)\\ Agent.Grab(novel)\\ Agent.Find(chair)\\ Agent.SitOn(chair)\\ Agent.Read(novel)"\}\end{tabular}                                                                                                                                                             \\
\midrule
\multicolumn{1}{l}{WebShop}                                                            & {\color[HTML]{FF0000} \{task\}}                                                                                                                                                                   & {\color[HTML]{FF0000} \{action\}}                                                                                                                                                                                                                                                                                                                                 & \begin{tabular}[c]{@{}l@{}}\{task: "Instruction: i'm looking to \\ buy a high resolution marine \\animal themed backdrop. the size \\should be 12x10ft, and price lower \\than 100.00 dollars\\ {[}button{]} Search {[}button\_{]}\\ Action:",\\ action: "\\ search{[}12x10ft high resolution \\marine animal backdrop{]}"\}\end{tabular}                          \\
\midrule
\textbf{Tabletop}                                                                      & {\color[HTML]{FF0000} \{task\}}                                                                                                                                                                   & {\color[HTML]{FF0000} \{action\}}                                                                                                                                                                                                                                                                                                                                 & \begin{tabular}[c]{@{}l@{}}\{task: "objects = {[}'yellow block', \\'green block', 'yellow bowl', \\'blue block', 'blue bowl', 'green bowl'{]}\\ \# move the green block to the \\top right corner.", \\ action: "\\corner\_pos = parse\_position(\\'top right corner')\\ put\_first\_on\_second(\\'green block', corner\_pos)"\}\end{tabular}                            \\
\bottomrule
\end{tabular}
\end{adjustbox}
\end{table}

\subsection{All-shot loss}
\begin{figure}[h]
\caption{We use all-shot loss for model alignment. We concatenate several examples into a single training sample and backpropagate through the loss on the blue actions in every example. There is no separator token between examples.
}
\centering
\vspace{-8pt}
\includegraphics[width=0.85\textwidth]{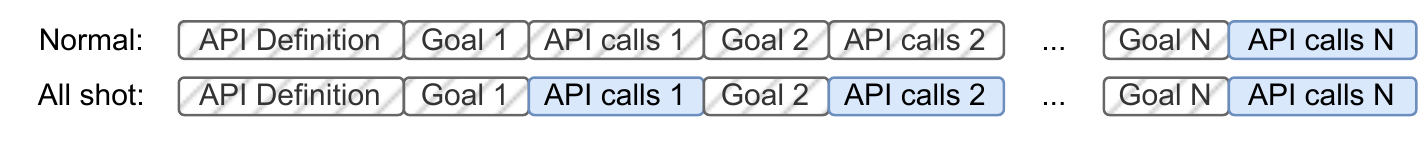}
\label{fig:all_shot}
\end{figure}
To construct the training samples, we concatenate API documents and multiple pairs of goal and API calls as one input sequence to the LLMs. We use an all-shot loss formulation illustrated in~\Cref{fig:all_shot} which learns to generate the API calls for every goal in a sequence.
We use this loss formulation because it empirically delivers better success rate, especially when using in-context demonstrations, than the conventional loss which only backpropagates the loss associated with the API calls for the last goal.

\subsection{Training details}

We finetune each model on the same dataset created with the method described in Section \ref{sec:training_data_section} for $8$ epochs. We use a max sequence length of $2048$ without packing and mix the data from all the tasks into a single dataset with random shuffling. In each sample, all the goal-action pairs are from the same task. We report the validation accuracy on the best checkpoint. We use a batch size of 16 and a constant learning rate of $1e-5$ for each model and train on an internal cluster of 4 A100 GPU's, each with 80GB RAM.

\begin{table}[]
\caption{The detailed performance on the \snact~of models with different techniques applied. Mean(standard deviation) values are provided for each task. There exists some inevitable randomness, but it won't cange the results by too much.}
\label{tab:baselines_over_techniques}
\begin{adjustbox}{max width=\textwidth}
\setlength{\tabcolsep}{2pt}
\begin{tabular}{@{}cccccccccc@{}}
\toprule
                                          & \textbf{Open}                           & \textbf{The Cat}                    & \textbf{Home}        & \textbf{Trip}        & \textbf{Google}      &                                        & \multicolumn{2}{c}{\textbf{WebShop}}        &                                     \\
\multirow{-2}{*}{\textbf{Task}}           & {\color[HTML]{1F1F1F} \textbf{Weather}} & {\color[HTML]{1F1F1F} \textbf{API}} & \textbf{Search}      & \textbf{Booking}     & \textbf{Sheets}      & \multirow{-2}{*}{\textbf{VirtualHome}} & \textbf{Long}        & \textbf{Short}       & \multirow{-2}{*}{\textbf{Tabletop}} \\
\midrule
\textit{\textbf{Zero-shot Baseline}}          & \multicolumn{1}{l}{} & \multicolumn{1}{l}{} & \multicolumn{1}{l}{} & \multicolumn{1}{l}{} & \multicolumn{1}{l}{} & \multicolumn{1}{l}{}  & \multicolumn{1}{l}{} & \multicolumn{1}{l}{} & \multicolumn{1}{l}{} \\
gpt4                                          & 81.3(1.7)            & 97.4(0.3)            & 76.6(1.1)            & 91.5(0.5)            & 5.7(0.0)             & 40.8(0.6) / 8.0(0.2)  & 0.0(0.0)             & -                    & -                    \\
llama-30b                                     & 39.0(0.0)            & 49.0(0.0)            & 0.0(0.0)             & 0.0(0.0)             & 0.0(0.0)             & 78.0(0.0) / 0.3(0.0)  & 0.0(0.0)             & -                    & -                    \\
starcoder                                     & 32.0(0.0)            & 71.0(0.0)            & 7.0(0.0)             & 13.3(0.0)            & 5.9(1.1)             & 22.0(0.0) / 3.7(0.0)  & 0.0(0.0)             & -                    & -                    \\
codegen-16B-mono                              & 7.0(0.0)             & 78.0(0.0)            & 0.0(0.0)             & 0.0(0.0)             & 1.4(0.0)             & 4.0(0.0) / 1.0(0.0)   & 0.0(0.0)             & -                    & -                    \\
\cmidrule(lr){1-10}
\textit{\textbf{Sys. Prompt}}                 & \multicolumn{1}{l}{} & \multicolumn{1}{l}{} & \multicolumn{1}{l}{} & \multicolumn{1}{l}{} & \multicolumn{1}{l}{} & \multicolumn{1}{l}{}  & \multicolumn{1}{l}{} & \multicolumn{1}{l}{} & \multicolumn{1}{l}{} \\
gpt4                                          & 78.4(0.3)            & 94.2(0.8)            & 72.7(2.0)            & 89.6(0.9)            & 28.6(0.0)            & 42.8(0.6) / 8.6(0.1)  & 0.0(0.0)             & -                    & -                    \\
llama-30b                                     & 50.0(0.0)            & 88.0(0.0)            & 0.0(0.0)             & 0.0(0.0)             & 11.4(0.0)            & 24.0(0.0) / 2.5(0.0)  & 0.0(0.0)             & -                    & -                    \\
starcoder                                     & 71.0(0.0)            & 91.0(0.0)            & 2.0(0.0)             & 7.5(0.0)             & 15.9(0.2)            & 26.0(0.0) / 4.9(0.0)  & 0.0(0.0)             & -                    & -                    \\
codegen-16B-mono                              & 32.0(0.0)            & 69.0(0.0)            & 0.0(0.0)             & 0.0(0.0)             & 7.1(0.0)             & 5.0(0.0) / 1.6(0.0)   & 0.0(0.0)             & -                    & -                    \\
\cmidrule(lr){1-10}
\textit{\textbf{3-shot}}                      & \multicolumn{1}{l}{} & \multicolumn{1}{l}{} & \multicolumn{1}{l}{} & \multicolumn{1}{l}{} & \multicolumn{1}{l}{} & \multicolumn{1}{l}{}  & \multicolumn{1}{l}{} & \multicolumn{1}{l}{} & \multicolumn{1}{l}{} \\
gpt4                                          & 93.0(0.0)             & 96.0(0.0)             & 97.0(0.0)             & 96.7(0.0)             & 62.9(0.0)            & 23.0(0.0) / 23.5(0.0)   & 0.0(0.0)             & 0.0(0.0)             & 81.0(0.0)             \\
llama-30b                                     & 78.0(0.0)            & 84.0(0.0)            & 66.0(0.0)            & 45.0(0.0)            & 37.1(0.0)            & 27.0(0.0) / 21.7(0.0) & 0.0(0.0)             & 30.6(0.0)            & 34.3(0.0)            \\
starcoder                                     & 91.0(0.0)            & 84.0(0.0)            & 82.0(0.0)            & 51.7(0.0)            & 48.0(1.1)            & 23.0(0.0) / 19.4(0.0) & 2.6(0.0)             & 0.0(0.0)             & 21.9(0.0)            \\
codegen-16B-mono                              & 63.7(0.5)            & 72.0(0.0)            & 52.0(0.0)            & 28.3(0.0)            & 31.5(0.5)            & 28.0(0.0) / 15.7(0.0) & 1.5(0.0)             & 6.6(0.0)             & 15.2(0.0)            \\
\cmidrule(lr){1-10}
\textit{\textbf{Alignment}}                   & \multicolumn{1}{l}{} & \multicolumn{1}{l}{} & \multicolumn{1}{l}{} & \multicolumn{1}{l}{} & \multicolumn{1}{l}{} & \multicolumn{1}{l}{}  & \multicolumn{1}{l}{} & \multicolumn{1}{l}{} & \multicolumn{1}{l}{} \\
llama-30b                                     & 100.0(0.0)           & 94.0(0.0)            & 85.0(0.0)            & 87.5(0.0)            & 4.3(0.0)             & 14.0(0.0) / 10.6(0.0) & 20.8(0.0)            & -                    & -                    \\
starcoder                                     & 95.0(0.0)            & 98.0(0.0)            & 78.0(0.0)            & 85.0(0.0)            & 10.0(0.0)            & 28.0(0.0) / 13.4(0.0) & 0.0(0.0)             & -                    & -                    \\
codegen-16B-mono                              & 99.0(0.0)            & 95.8(0.6)            & 78.0(0.0)            & 73.3(0.0)            & 10.0(0.0)            & 10.0(0.0) / 11.5(0.0) & 30.3(0.0)            & -                    & -                    \\
\cmidrule(lr){1-10}
\textit{\textbf{Sys. Prompt + 3-shot}}        & \multicolumn{1}{l}{} & \multicolumn{1}{l}{} & \multicolumn{1}{l}{} & \multicolumn{1}{l}{} & \multicolumn{1}{l}{} & \multicolumn{1}{l}{}  & \multicolumn{1}{l}{} & \multicolumn{1}{l}{} & -                    \\
gpt4                                          & 99.0(0.0)            & 98.0(0.0)            & 98.0(0.0)            & 99.2(0.0)            & 68.6(0.0)            & 29.0(0.0) / 21.7(0.0) & 0.0(0.0)             & 0.0(0.0)             & 83.8(0.0)            \\
llama-30b                                     & 66.0(0.0)            & 82.0(0.0)            & 63.0(0.0)            & 45.8(0.0)            & 27.1(0.0)            & 34.0(0.0) / 20.5(0.0) & 0.0(0.0)             & 0.0(0.0)             & 34.6(0.2)            \\
starcoder                                     & 92.0(0.0)            & 91.0(0.0)            & 73.0(0.0)            & 54.2(0.0)            & 50.0(0.2)            & 28.0(0.0) / 15.0(0.0) & 0.0(0.0)             & 0.0(0.0)             & 23.4(0.3)            \\
codegen-16B-mono                              & 64.2(0.3)            & 70.0(0.0)            & 45.0(0.0)            & 22.5(0.0)            & 28.6(0.9)            & 27.0(0.0) / 15.7(0.0) & 0.0(0.0)             & 0.0(0.0)             & 14.6(0.2)            \\
\cmidrule(lr){1-10}
\textit{\textbf{Sys. Prompt + Alignment}}     &                      & \multicolumn{1}{l}{} & \multicolumn{1}{l}{} & \multicolumn{1}{l}{} & \multicolumn{1}{l}{} & \multicolumn{1}{l}{}  & \multicolumn{1}{l}{} & \multicolumn{1}{l}{} & \multicolumn{1}{l}{} \\
llama-30b                                     & 100.0(0.0)           & 94.0(0.0)            & 79.0(0.0)            & 80.8(0.0)            & 5.7(0.0)             & 10.0(0.0) / 10.3(0.0) & 0.6(0.0)             & -                    & -                    \\
starcoder                                     & 98.7(0.2)            & 97.0(0.0)            & 79.0(0.0)            & 84.2(0.0)            & 10.0(0.0)            & 18.0(0.0) / 10.3(0.0) & 0.0(0.0)             & -                    & -                    \\
codegen-16B-mono                              & 99.0(0.0)            & 96.0(0.0)            & 77.0(0.0)            & 75.8(0.0)            & 8.6(0.0)             & 7.0(0.0) / 10.0(0.0)  & 25.7(0.0)            & -                    & -                    \\
\cmidrule(lr){1-10}
\textit{\textbf{3-shot + Alignment}}          & \multicolumn{1}{l}{} & \multicolumn{1}{l}{} & \multicolumn{1}{l}{} & \multicolumn{1}{l}{} & \multicolumn{1}{l}{} & \multicolumn{1}{l}{}  & \multicolumn{1}{l}{} & \multicolumn{1}{l}{} & \multicolumn{1}{l}{} \\
llama-30b                                     & 100.0(0.0)           & 94.0(0.0)            & 88.0(0.0)            & 89.2(0.0)            & 4.3(0.0)             & 20.0(0.0) / 26.3(0.0) & 19.5(0.0)            & 15.1(0.0)            & 6.9(0.2)             \\
starcoder                                     & 100.0(0.0)           & 96.0(0.0)            & 91.0(0.0)            & 84.2(0.0)            & 15.7(0.0)            & 48.0(0.0) / 21.3(0.0) & 0.0(0.0)             & 0.0(0.0)             & 13.9(0.3)            \\
codegen-16B-mono                              & 99.0(0.0)            & 97.9(0.2)            & 80.0(0.0)            & 77.5(0.0)            & 16.4(0.9)            & 38.0(0.0) / 18.6(0.0) & 6.5(0.0)             & 17.5(0.0)            & 16.2(0.0)            \\
\cmidrule(lr){1-10}
\textit{\textbf{Prompt + 3-shot + Alignment}} & \multicolumn{1}{l}{} & \multicolumn{1}{l}{} & \multicolumn{1}{l}{} & \multicolumn{1}{l}{} & \multicolumn{1}{l}{} & \multicolumn{1}{l}{}  & \multicolumn{1}{l}{} & \multicolumn{1}{l}{} & \multicolumn{1}{l}{} \\
llama-30b                                     & 100.0(0.0)           & 94.0(0.0)            & 87.0(0.0)            & 85.8(0.0)            & 2.9(0.0)             & 16.0(0.0) / 24.3(0.0) & 0.0(0.0)             & 0.0(0.0)             & 7.5(0.1)             \\
starcoder                                     & 99.0(0.0)            & 97.0(0.0)            & 83.0(0.0)            & 80.8(0.0)            & 21.2(0.3)            & 31.0(0.0) / 18.4(0.0) & 0.0(0.0)             & 0.0(0.0)             & 13.9(0.3)            \\
codegen-16B-mono                              & 97.7(0.2)            & 99.0(0.0)            & 82.0(0.0)            & 77.5(0.0)            & 19.8(0.3)            & 29.0(0.0) / 17.2(0.0) & 0.0(0.0)             & 3.5(0.0)             & 16.2(0.0)                    \\
\bottomrule
\end{tabular}
\end{adjustbox}
\end{table}

\subsection{Extended results for \Cref{sec:experiments}}
We list out the detailed results of \Cref{sec:experiments} in \Cref{tab:baselines_over_techniques}, where we report the model performance on all the possible combinations of the three proposed techniques. The main observations are all covered in \Cref{sec:experiments}. We run each job 3 times, and report the mean and standard deviation of the main metrics. Their are some inevitable randomness happens in API or example retrieval, public API services and the environment provided in Webshop and Tabletop. Even though randoness exists, we observe that they barely change the final results. Thus, we only report the mean value everywhere else in the paper.

\section{API Selection Complexity Score}
\label{sec:app_complexity_metrics}
\subsection{Complexity score}
This section introduces a complexity score system designed to measure the intrinsic complexity and difficulty of the tasks from \textit{ToolBench}. The complexity score system aims to provide a quantitative measure of the intrinsic complexity of the tests given the examples by calculating the probability of the tests being derived or converted from the examples; and the derivation or conversion is performed in a random system with all possible outcomes equally likely. This score serves to assess the inherent level of difficulty involved in transitioning from one scenario to another, thereby assisting researchers and developers in benchmark evaluation and analysis.


\subsubsection{The likelihood of a test being derived from an example}
In the complexity score system proposed herein, the calculation of the complexity score involves assessing the probability or likelihood of the tests being derived from an example in the particular task. Given a demonstration example $e$ and a set of API functions $\mathcal{D}$, the derivation of a particular test sample $t$ involves two major steps: 1) remove all the unused API calls while keeping all the necessary ones and 2) insert the new API calls that $e$ does not cover. Given a random system, where all possible outcomes are equally likely, we suppose the deletion possibility of each API call from $e$ is 50\%, while the insertion possibilities of the correct API call is $1 / |\mathcal{D}|$, where $|\mathcal{D}|$ is the total number of API functions of the given task. If $t$ or $e$ contains multiple calls to the same API function, we consider them as different API calls, because they are usually not interchangeable. Based on these assumptions, the likelihood of generating a test sample $t$ is calculated using Equation (1). 


\begin{equation}
    \label{eq:equation1}
    p(t~|~e, \mathcal{D})=\left(\frac{1}{2}\right)^{|e|} \left(\frac{1}{|\mathcal{D}|}\right)^{|t \setminus e|}
\end{equation}

where $|e|$ represents the number of API calls in the example $e$, and $|t \setminus e|$ is the number of uncovered API calls in the test sample. Suppose we have a task that has 10 API functions in total $\{a_i\}_1^{10}$, and the demonstration example covers $\{a_1, a_2, a_3, a_4\}$, but the test sample requires $\{a_1, a_2, a_6, a_4, a_5\}$. In the first step, the probability of successfully dropping $a_3$ while keeping the rest ones in $e$ is $\left(\frac{1}{2}\right)^4$. Then, the probability of correctly adding in the uncovered ones, $a_5$ and $a_6$, is $\left(\frac{1}{10}\right)^2$. Note that we do not take the order of API calls into consideration for the purpose of being simple without losing generosity.

\subsubsection{The distance between a test and example pair}
We first define the distance $d$ between one particular test and example pair by take the logarithm of the reciprocal of Equation (1) as:
\begin{equation}
    \label{eq:equation2}
    d(t, e) = \log\left[\frac{1}{p(t~|~e, \mathcal{D})}\right]
\end{equation}
The use of the reciprocal in the expression aligns the complexity score with the definition of complexity, where a higher score indicates a greater level of complexity. Additionally, applying the logarithm to the reciprocal value aids in addressing the magnitude gap. The logarithm function compresses the range of values, reducing the impact of extreme values and creating a more manageable scale. This normalization ensures that the complexity score is not disproportionately influenced by outliers or extreme values, providing a more balanced representation of complexity across the range of input values. By combining the reciprocal and logarithm, the expression effectively balances the score by aligning it with the definition of complexity and mitigating the impact of magnitude differences in the input values.

\subsubsection{Complexity score of a task}
\label{sec:complexity}
Based on the complexity score of generating a test from an example, we can construct the complexity score $S$ of a given task. The score $S = f(\mathcal{T}, \mathcal{X}, \mathcal{D})$ is a function of the test samples $\mathcal{T}$, the demonstration examples $\mathcal{X}$ and the API functions $\mathcal{D}$ of each task. 

\begin{equation}
\begin{aligned}
    S(\mathcal{T}, \mathcal{X}, \mathcal{D}) &= \mathbb{E}_{t \in \mathcal{T}}\min\nolimits_{e \in \mathcal{X}} d(t, e) \\
    &= \mathbb{E}_{t \in \mathcal{T}}\min\nolimits_{e \in \mathcal{X}}\log\left[\frac{1}{p(t~|~e, \mathcal{D})}\right]  \\
    &= - \mathbb{E}_{t \in \mathcal{T}}\max\nolimits_{e \in \mathcal{X}}\log\left[\left(\frac{1}{2}\right)^{|e|} \left(\frac{1}{|\mathcal{D}|}\right)^{|t \setminus e|}\right]
\end{aligned}
\end{equation}

This score ranges from zero to infinity. The larger the score is, the more challenging a task is in terms of API selection. We calculate this score for both the original \snact~(\Cref{tab:all_tasks}) and the training data we created for alignment \Cref{tab:training_data}. They share the same $\mathcal{D}$ and $\mathcal{T}$, but have a different $\mathcal{X}$, so that their API selection complexities are different for each task.



\subsection{Complexity score on the \snact}
In this section we demonstrate how the complexity score behaves on the \snact. 

\subsubsection{Computation details}
For the Trip Booking, Home Search, Virtual Home, and Google Sheets tasks, the set of API functions $\mathcal{D}$ is the same as described in~\cref{sec:benchmark_details}. For the single-step, single-API-call tasks, Open Weather and The Cat API, each valid URL with parameters is treated as a unique API option in set $\mathcal{D}$. In total, Open Weather has 37 API options, while The Cat API has 52 API options. In the case of the Tabletop task, since there are no predefined correct answers for the test cases, we divide the three set of "Tabletop Manipulation" examples\footnote{https://code-as-policies.github.io/} into 65 single-step samples. Note that for the WebShop task, since there are only two API functions always covered by the example set, the complexity score is 0 by definition. 

\subsubsection{Reversed-F1 Score}
For comparison purpose, we also consider the simple Reversed-F1 (r-F1) distance $d_{r-F1}$, derived from the conventional F1 score\cite{davis2006relationship}, between one particular test and example pair as

\begin{equation}
    d_{r-F1}(t, e) = (1-F1(t,e)) *100
\end{equation}
We multiply 100 to the score to align with the range of the complexity score defined above. Follow the same definition proposed in \cref{sec:complexity}, we can construct the r-F1 score $S_{r-F1}$ of a given task as:

\begin{equation}
\begin{aligned}
    S_{r-F1}(\mathcal{T}, \mathcal{X}) &= \mathbb{E}_{t \in \mathcal{T}}\min\nolimits_{e \in \mathcal{X}} d_{r-F1}(t, e) \\
    &= \mathbb{E}_{t \in \mathcal{T}}\min\nolimits_{e \in \mathcal{X}}\left[(1-F1(t,e)) *100\right]
\end{aligned}
\end{equation}

\subsubsection{Measurements}

\begin{figure}
    \caption{Spearman's correlation coefficient(SCC) is computed separately for two comparisons: (1) complexity score and error rate, and (2) reversed F1 score and error rate on five tasks: (1) Open Weather, (2) The Cat API, (3) Home Search, (4) Trip Booking, and (5) Virtual Home. }\label{fig:scc}
    \includegraphics[width=\textwidth]{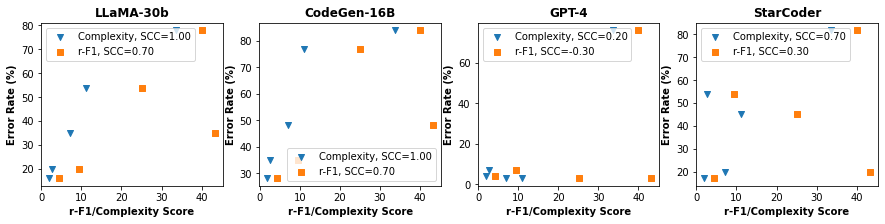}
\end{figure}

\begin{wraptable}[7]{r}{0.63\textwidth}
\vspace{-12pt}
    \centering
    \caption{Spearman’s Correlation Coefficients}
    \begin{tabular}{ccccc}
        \toprule
        & GPT-4 & LLaMA & CodeGen & StarCoder  \\
        \midrule
        Complexity & 0.2 & 1.0 & 1.0 & 0.7 \\
        r-F1 & -0.3 & 0.7 & 0.7 & 0.3 \\
        \bottomrule
    \end{tabular}
    \label{table:scc}
\end{wraptable}
In this section, Spearman's Correlation Coefficient (SCC) \cite{hauke2011comparison} is employed to assess the effectiveness of the proposed complexity score. The evaluation involves the analysis of five different tasks using three models: GPT-4, LLaMA-30b, CodeGen-16b, and StarCoder. We only include the five tasks without advanced reasoning from \cref{tab:all_tasks}, as the advanced reasoning breaks the correlation between the API selection difficulty and the final model performance. The complexity score and the r-F1 score are calculated for each task. SCC is then computed separately for two comparisons: (1) complexity score and error rate, and (2) reversed F1 score and error rate, for all five tasks. The results are illustrated in \cref{fig:scc} and \cref{table:scc}.

The findings of the study reveal near-perfect Spearman's correlation coefficient (SCC) between the complexity score and the error rate for the LLaMA-30b, CodeGen-16b and StarCoder models. This strong correlation indicates that the proposed complexity score system accurately captures the intrinsic difficulty of these tasks.

For more powerful models like GPT4, which exhibit near-perfect accuracy (above 93\%) for low-complexity tasks (complexity < 12) such as Open Weather, The Cat API, Home Search, and Trip Booking, the SCC becomes relatively sensitive to any randomness or turbulence during the experiments. Consequently, the complexity score system shows a non-perfect SCC of 0.2 in this case.

Despite the sensitivity of the SCC in the GPT4 experiments, the complexity score remains a superior indicator of task difficulty compared to the r-F1 score. It effectively captures the inherent difficulty of each task and provides valuable insights into task complexity. Overall, complexity score is more effective at capturing the inherent difficulty of each task, thus providing valuable insights into task complexity.

The obtained results provide empirical evidence supporting the validity and reliability of the proposed complexity score system. The high SCC values signify a consistent relationship between the complexity score and the error rate across different models and tasks. This correlation strengthens the argument that the complexity score accurately captures the complexity and difficulty of the benchmarks, enabling researchers and developers to assess and compare the inherent challenges associated with different tasks.



\end{document}